\documentclass[10pt,twocolumn,letterpaper]{article}

\usepackage[pagenumbers]{wacv} % To force page numbers, e.g. for an arXiv version

\usepackage{graphicx}
\usepackage{amsmath}
\usepackage{amssymb}
\usepackage{booktabs}
\usepackage{multirow}
\usepackage[acronym]{glossaries}
\usepackage{float}
\usepackage{amsthm}
\usepackage{overpic}
\usepackage{colortbl,xcolor}
\usepackage[accsupp]{axessibility} % Improves PDF readability for those with disabilities.
\usepackage{makecell}

\usepackage[pagebackref,breaklinks,colorlinks]{hyperref}

\usepackage[capitalize]{cleveref}
\crefname{section}{Sec.}{Secs.}
\Crefname{section}{Section}{Sections}
\Crefname{table}{Table}{Tables}
\crefname{table}{Tab.}{Tabs.}
\Crefname{subfigure}{Figure}{Figures}
\crefname{subfigure}{Fig.}{Figs.}

\def\wacvPaperID{167} % *** Enter the WACV Paper ID here
\def\confName{WACV}
\def\confYear{2025}

\newcommand{\std}[1]{{\tiny$\pm$#1}}

\newtheorem{definition}{Definition}

\newacronym{dg}{DG}{domain generalization}
\newacronym{as}{AS}{abdominal segmentation}
\newacronym{lss}{LSS}{lumbar spine segmentation}
\newacronym{ls}{LS}{lung segmentation}
\newacronym{ub}{UB}{upper bound} 
\newacronym{pp}{pp}{percentage points} 
\newacronym{uda}{UDA}{unsupervised domain adaptation}
\newacronym{sdg}{SDG}{\emph{single-source} DG}
\newacronym{mdg}{MDG}{\emph{multi-source} DG}
\newacronym{scm}{SCM}{structural causal model}
\newacronym{kl}{KL-Div}{Kullback–Leibler divergence}
\newacronym{sd}{SD}{Stable Diffusion}
\newacronym[plural=LDMs,firstplural=latent DMs (LDMs)]{ldm}{LDM}{latent DM}
\newacronym{vae}{VAE}{Variational Auto-Encoder}
\newacronym{iid}{i.i.d.}{independent and identically distributed }
\newacronym{irm}{IRM}{Invariant Risk Minimization}
\newacronym{gin}{GIN}{global intensity non-linear augmentation}
\newacronym{ipa}{IPA}{interventional pseudo-correlation augmentation}
\newacronym{ood}{OOD}{out-of-distribution}
\newacronym[plural=NNs,firstplural=neural networks (NNs)]{nn}{NN}{neural network}
\newacronym[plural=DMs,firstplural=diffusion models (DMs)]{dm}{DM}{diffusion model}

\makeglossaries

\begin{document}

\title{Generalizable Single-Source Cross-modality Medical Image Segmentation via Invariant Causal Mechanisms}

\author{
Boqi Chen$^{1,2,3}$
\qquad
Yuanzhi Zhu$^{3}$
\qquad
Yunke Ao$^{1,2,4}$
\qquad
Sebastiano Caprara$^{4,5}$\\
Reto Sutter$^{4,5}$
\qquad
Gunnar Rätsch$^{1,2}$
\qquad
Ender Konukoglu$^{3}$
\qquad
Anna Susmelj$^{2,3}$
 \\
$^{1}$ Department of Computer Science, ETH Zurich \quad $^{2}$ ETH AI Center \\ \quad $^{3}$ Computer Vision Lab, ETH Zurich \quad $^{4}$ Balgrist University Hospital \quad $^{5}$ University of Zurich
}
\maketitle

%%%%%%%%% ABSTRACT
\begin{abstract}
Single-source domain generalization (SDG) aims to learn a model from a single source domain that can generalize well on unseen target domains. This is an important task in computer vision, particularly relevant to medical imaging where domain shifts are common. In this work, we consider a challenging yet practical setting: SDG for cross-modality medical image segmentation. We combine causality-inspired theoretical insights on learning domain-invariant representations with recent advancements in diffusion-based augmentation to improve generalization across diverse imaging modalities. Guided by the ``intervention-augmentation equivariant'' principle, we use controlled diffusion models (DMs) to simulate diverse imaging styles while preserving the content, leveraging rich generative priors in large-scale pretrained DMs to comprehensively perturb the multidimensional style variable. 
%The rich generative priors in large-scale pretrained DMs ensure comprehensive perturbation of the multidimensional style variable, surpassing conventional augmentation techniques.  
Extensive experiments on challenging cross-modality segmentation tasks demonstrate that our approach consistently outperforms state-of-the-art SDG methods across three distinct anatomies and imaging modalities. The source code is available at \href{https://github.com/ratschlab/ICMSeg}{https://github.com/ratschlab/ICMSeg}. 
% Extensive experiments on two challenging cross-modality segmentation tasks show that our approach outperforms state-of-the-art SDG methods by a significant margin. 
% Code will be made available upon acceptance.
% This work provides a practical solution for enhancing the generalizability of medical image segmentation models trained on a single domain, addressing the critical need for robust performance across varied imaging modalities in clinical settings.
\end{abstract}

%%%%%%%%% BODY TEXT
\section{Introduction}
\label{sec:intro}

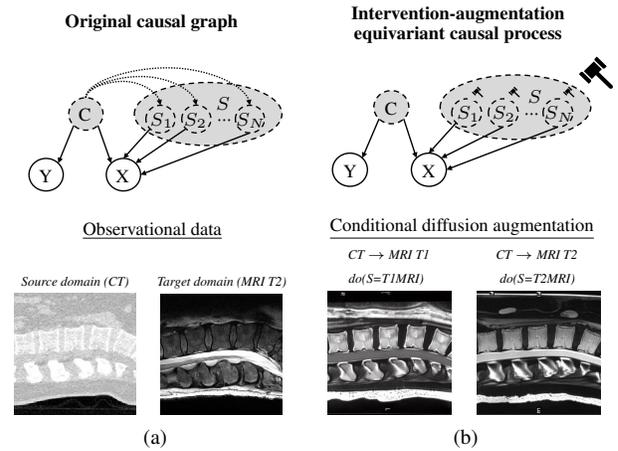
\begin{figure}[t]
    \centering
    \begin{subfigure}{0.49\linewidth}
        \centering
        \begin{overpic}[width=1 \columnwidth]{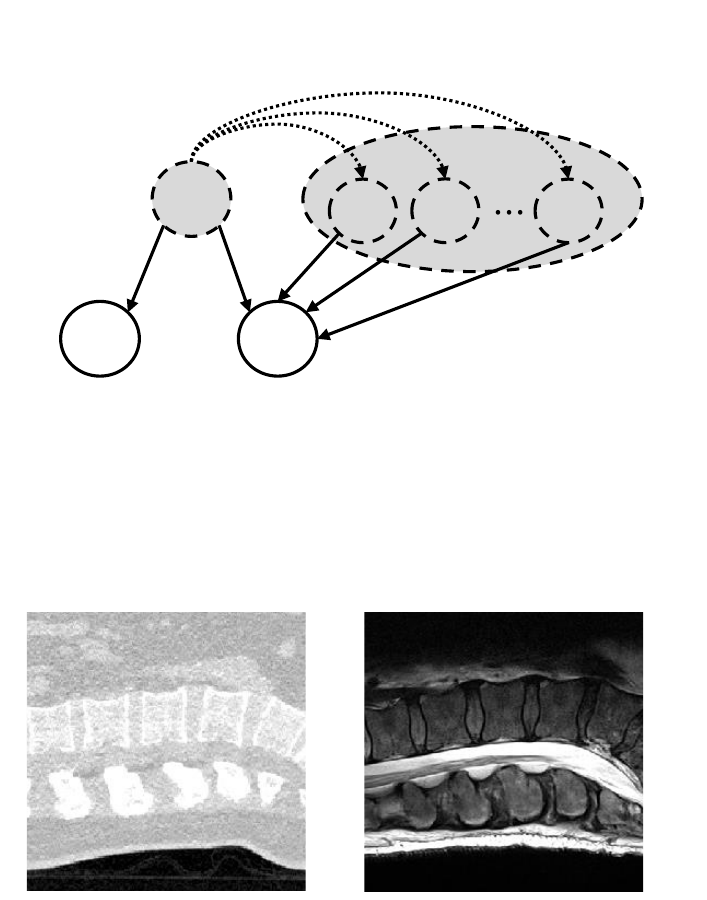}
        \put(16., 100){\color{black}\scriptsize{\textbf{Original causal graph}}}
        \put(19.4, 76.2){\color{black}\scriptsize{C}}
        \put(54.5, 79.){\color{black}\scriptsize{$S$}}
        \put(37.7, 75){\color{black}\scriptsize{$S_1$}}
        \put(46.3, 75){\color{black}\scriptsize{$S_2$}}
        \put(59.6, 75){\color{black}\scriptsize{$S_N$}}
        \put(9.3, 60.5){\color{black}\scriptsize{Y}}
        \put(28.9, 60.5){\color{black}\scriptsize{X}}
        \put(20.5, 47){\color{black}\scriptsize\underline{Observational data}}
        \put(4.8, 34){\color{black}\tiny{\textit{Source domain (CT)}}}
        \put(39.5, 34){\color{black}\tiny{\textit{Target domain (MRI T2)}}}
        \end{overpic}
        \caption{}
        \label{fig1_a}
    \end{subfigure}
    % \hfill
    \begin{subfigure}{0.48\linewidth}
        \centering
        \begin{overpic}[width=1 \columnwidth]{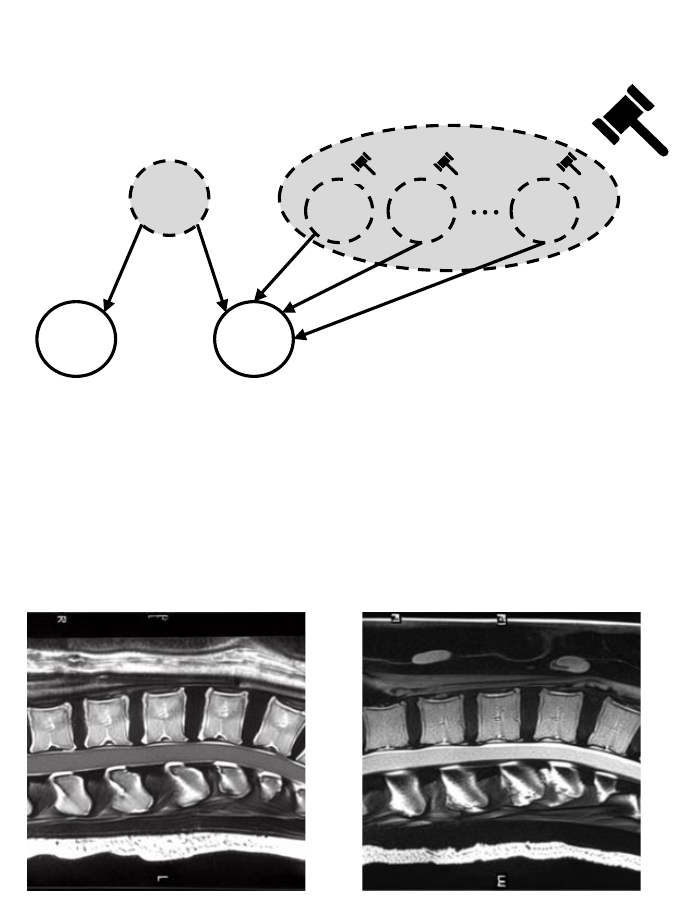}
        \put(8.8, 100){\color{black}\scriptsize{\textbf{Intervention-augmentation }}}
        \put(9.5, 95){\color{black}\scriptsize{\textbf{equivariant causal process}}}
        \put(17.2, 76.2){\color{black}\scriptsize{C}}
        \put(53, 79.){\color{black}\scriptsize{$S$}}
        \put(35.2, 75){\color{black}\scriptsize{$S_1$}}
        \put(44.1, 75){\color{black}\scriptsize{$S_2$}}
        \put(57.1, 75){\color{black}\scriptsize{$S_N$}}
        \put(6.8, 60.5){\color{black}\scriptsize{Y}}
        \put(26.5, 60.5){\color{black}\scriptsize{X}}
        \put(3.5, 47){\color{black}\scriptsize\underline{Conditional diffusion augmentation}}
        \put(8, 40){\color{black}\tiny{\textit{CT $\rightarrow$ MRI T1}}}
        \put(45, 40){\color{black}\tiny{\textit{CT $\rightarrow$ MRI T2}}}
        \put(8.2, 34){\color{black}\tiny{\textit{do(S=T1MRI)}}}
        \put(46, 34){\color{black}\tiny{\textit{do(S=T2MRI)}}}
        \end{overpic}
        \caption{}
        \label{fig1_b}
    \end{subfigure}
    \caption{ Overview of causal framework. (a) SCM for data generative process and examples of samples from the observational distribution. (b) Causal graph after \textit{do}-intervention on style variables and examples of corresponding equivariant augmentations generated with the conditional diffusion model.}
    \label{fig1}
    % \vspace{-0.3cm}
\end{figure}

Deep convolutional \glspl{nn} have recently achieved remarkable performance in medical image segmentation~\cite{ronneberger2015u, isensee2018nnu, zhou2018unet++, antonelli2022medical, li2018h, milletari2016v}. Despite their success, these models' performance heavily relies on the assumption that training and testing samples are \gls{iid}. In medical applications, however, distribution shifts between source (training) and target (test) data frequently occur due to varying scanning protocols, different device vendors, and diverse imaging modalities, etc. Directly applying models trained on a source domain to an unseen target domain with a different distribution typically results in significant performance degradation~\cite{kamnitsas2017unsupervised}. Broadly, two main types of domain shift exist in medical imaging: cross-center, where the shift primarily stems from different acquisition protocols and devices, and cross-modality, where the shift is caused by different imaging modalities (\eg, CT to MRI). The latter presents a more substantial distribution shift between the source and target domains.
% Numerous efforts have addressed the domain shift problem, with \gls{uda} being the most popular approach. 

To address the domain shift problem, researchers have proposed two primary solutions: \gls{uda}~\cite{ganin2015unsupervised, murez2018image, zhang2018fully, hoffman2018cycada, chen2019crdoco} and \gls{dg}~\cite{du2020learning, motiian2017unified, muandet2013domain, mahajan2021domain, nguyen2023causal, lv2022causality, chen2023meta, ouyang2022causality}. \gls{uda} assumes access to unlabeled data from the target domain, enabling joint training on both source and target images to learn domain-invariant features. While effective for known target domains, \gls{uda} cannot guarantee high performance in unseen domains and often faces challenges in accessing target domain data due to privacy regulations. In contrast, \gls{dg} aims to train a generalizable and robust predictive model using only source domain data, thus overcoming the limitations of \gls{uda} and providing a compelling alternative for real-world medical applications. Conventional \gls{dg}~\cite{du2020learning, motiian2017unified, muandet2013domain} approaches seek to learn a domain-invariant feature space by aligning distributions among multiple source domains. Recent works~\cite{arjovsky2019invariant, mahajan2021domain} have established a more formal link between learned domain-invariances and the data generative process from a causal perspective, leading to improved performance in \gls{mdg}. However, acquiring multi-domain training data for medical applications remains challenging due to privacy issues and data scarcity. A more practical alternative is to train a model on a single source domain and enhance its generalizability to unseen target domains, known as \gls{sdg}. Adapting \gls{mdg} strategies to \gls{sdg}, however, is challenging due to the limited diversity in training data.

In this paper, we consider a challenging yet practical setting: cross-modality \gls{sdg}. While previous studies have explored this problem in medical imaging, most rely on basic augmentation techniques, such as manually-selected image-level transformations~\cite{zhou2022dn, ouyang2022causality} or random intensity adjustments~\cite{xu2020robust, su2023rethinking}, which are often insufficient for bridging domain gaps in cross-modality contexts. 
% In this paper, we consider a challenging yet practical setting: cross-modality \gls{sdg}. While several studies have attempted to tackle this problem in medical imaging, they primarily rely on generic augmentation techniques such as manually-selected image-level transformations~\cite{zhou2022dn, ouyang2022causality} or random intensity changes~\cite{xu2020robust, su2023rethinking}. Such augmentations are insufficient for reducing domain gaps, particularly when confronted with large distribution shifts in cross-modality scenarios.
Recently, diffusion-based augmentation methods have shown significant promise in handling large distribution shifts in the natural image domain~\cite{gong2023prompting, jia2023dginstyle}. This success can be attributed to two key factors: (1) the rich generative prior from large-scale pretraining and (2) comprehensive style mining through text prompting, resulting in refined, target-specific augmentations that improve generalization. However, their efficacy in medical imaging remains unexplored.
% Recently, diffusion-based augmentation methods have demonstrated remarkable efficacy in handling strong distribution shifts within the natural image domain~\cite{gong2023prompting, jia2023dginstyle}. The success can be attributed to two key factors: (1) the rich generative prior from large-scale pretraining, and (2) comprehensive style mining via text prompting, which leads to more refined and target-specific augmentations, ultimately enhancing generalization performance. However, their efficacy in medical imaging remains unexplored.

Motivated by this, we propose to combine the diffusion-based augmentations with recent theoretical advancements in causal identifiability~\cite{von2021self} and generalization via invariant mechanisms~\cite{arjovsky2019invariant}. Specifically, we assume a \gls{scm} for the data generation process, representing the dependency among observed data, labels (\ie, segmentation masks), and unobserved latent variables of content and style (\cref{fig1_a}). We then develop a style-augmentation method based on the ``intervention-augmentation equivariant'' principle~\cite{ilse2021selecting} using controlled \glspl{dm}, which enables simulating diverse imaging styles across modalities via text prompting, while preserving the content. We posit that the style variable is inherently multidimensional, and our strong augmentation strategy comprehensively intervenes on it (\cref{fig1_b}), whereas previous causality-inspired \gls{sdg} methods~\cite{chen2023meta, ouyang2022causality} rely on basic image-level transformations that may not fully disentangle style from content, potentially limiting model generalizability to unseen domains.
% Our approach contrasts with previous causality-inspired \gls{sdg} methods~\cite{chen2023meta, ouyang2022causality}, which rely on basic image-level transformations that may not fully disentangle style from content, potentially compromising generalization to unseen domains. 
% Our approach is generic to various segmentation network architectures.

In summary, the main contributions of this paper are summarized as:
\begin{itemize}
    \item We leverage recent theoretical advances in \gls{dg} under causal assumptions to design our approach for learning domain-invariant representations.
    \item We propose a style intervention method that utilizes rich generative prior of pretrained text-to-image \glspl{dm}.
    \item We demonstrate our method's effectiveness through extensive cross-modality medical image segmentation experiments, {consistently outperforming} state-of-the-art \gls{sdg} methods {across three distinct anatomies and imaging modalities}.
\end{itemize}

%-------------------------------------------------------------------------
\section{Related Work}
\label{sec:related_work}
\subsection{Single-source Domain Generalization}

\gls{sdg} is a more challenging and under-explored task compared to \gls{mdg}. Due to insufficient diversity of the training data, \gls{sdg} approaches usually first expand the distribution of the source domain via data augmentation techniques such as manually-selected image-level transformations~\cite{zhang2020generalizing, romera2018train, devries2017improved} and learning-based augmentation~\cite{qiao2020learning, li2021progressive, wang2021learning, zhao2020maximum}. 
% due to insufficient diversity in the training data. While \gls{mdg} allows for learning domain-invariant feature representations from multiple source domains, \gls{sdg} approaches typically tackle this limitation by expanding the distribution of the source domain via data augmentation techniques. These techniques include manually selected image-level transformations~\cite{zhang2020generalizing, romera2018train, devries2017improved} and learning-based augmentation~\cite{qiao2020learning, li2021progressive, wang2021learning, zhao2020maximum}. 
For instance, 
% Qiao et al.~\cite{qiao2020learning} introduce meta-learning based adversarial domain augmentation using a Wasserstein Auto-Encoder. 
in natural image domain, Li et al.~\cite{li2021progressive} leverage random style transfer with adaptive instances normalization~\cite{huang2017arbitrary}, followed by contrastive learning. Chen et al.~\cite{chen2023meta} apply predefined image-level transformations such as brightness and contrast adjustments to create auxiliary domains and then conduct counterfactual inference to learn robust features by analyzing the causal effect of simulated domain shifts. {While effective for natural images, such augmentations do not introduce sufficient variability for medical imaging data, which contains highly specialized and more nuanced visual patterns.}

To this end, several approaches tailored to medical imaging have been proposed. Zhang et al.~\cite{zhang2020generalizing} utilize stacked photometric and geometric transformations to simulate changes in image quality, appearance, and spatial configuration. Zhou et al.~\cite{zhou2022dn} employed Bézier transformations with switchable batch normalization layers based on style proximity. Ouyang et al.~\cite{ouyang2022causality} introduce local variability through randomly-sampled kernel augmentations with pseudo-correlation map-based mixing, simulating more diverse domain shifts. Su et al.~\cite{su2023rethinking} further advance this concept by integrating global and local Bézier transformations with saliency-based fusion, potentially offering more semantically meaningful augmentations. Despite these advancements, these methods are based on generic image-level manipulations, limiting their ability to capture complex domain variations. In contrast, \glspl{dm} leverage rich generative priors and enable target-specific augmentations through text prompting, introducing more complex and comprehensive distribution shifts.

\subsection{Causality for Domain Generalization}
The theory of causation has been extensively explored as a foundation for \gls{dg}~\cite{scholkopf2012causal, learningdataset, zhang2020causal, scholkopf2021toward}.  The idea of invariance is closely linked to causality~\cite{muandet2013domain}. Peters et al.~\cite{peters2016causal} introduce the concept of invariant causal prediction, identifying causal relationships that are stable across different environments. Arjovsky et al. \cite{arjovsky2019invariant} establish a formal link among generalization, causation and invariance. These theoretical advancements in understanding the relationship between causation and invariances have sparked a distinct line of research~\cite{lv2022causality, ouyang2022causality, mitrovic2020representation, mahajan2021domain, chen2023meta} focused on designing loss functions and regularization techniques based on these principles. 

\section{Method}
\label{sec:method}
In this section, we first present a {motivation} that obtaining domain-invariant representations using the direct causal parents of segmentation labels in the corresponding \gls{scm} leads to an optimal solution for the \gls{dg} problem. We demonstrate how to achieve such an optimal solution in \gls{sdg} using the ``soft'' intervention with contrastive learning. We then introduce a diffusion-based augmentation technique that is based on the ``intervention-augmentation equivariance'' principle such that it can serve as a surrogate tool for simulating interventions on unobserved variables in the \gls{scm}. Finally, we provide an overview of the two training stages of our method. 

\begin{figure*}[t]
    \centering
    \begin{subfigure}{0.95\linewidth}
        \centering
        \begin{overpic}[width=1 \columnwidth]{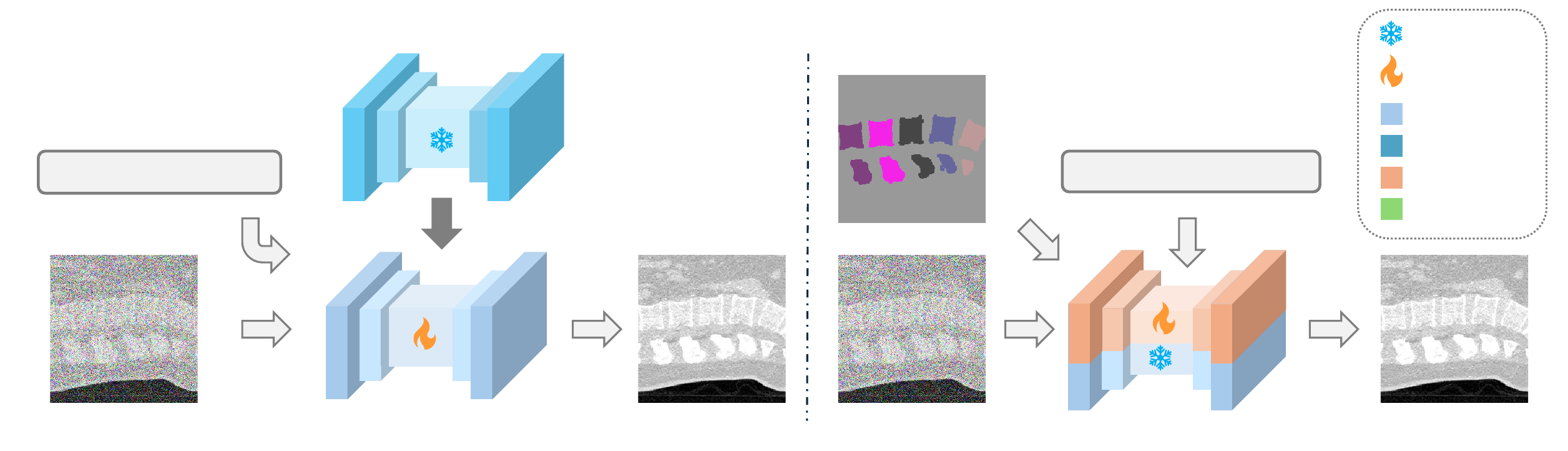}
        \put(3.3, 17.3){\color{black}\scriptsize{{\textit{“sagittal lumber spine”}}}}
        \put(3.4, 19.8){\color{black}\scriptsize{{\textbf{Style-agnostic prompt}}}}
        \put(3.3, 1.3){\color{black}\scriptsize{{Noisy image $\mathbf{x}_t$}}}
        \put(30., 14){\color{black}\scriptsize{{initialize}}}
        % \put(29., 25.5){\color{black}\scriptsize{$U^B$}}
        % \put(26., 2){\color{black}\scriptsize{$U^{D_0}$}}
        \put(40.6, 1.4){\color{black}\scriptsize{{Clean image $\mathbf{x}_0$}}}
        \put(53.6, 1.4){\color{black}\scriptsize{{Noisy image $\mathbf{x}_t$}}}\
        \put(52., 13.2){\color{black}\scriptsize{{Conditional image $c_i$}}}
        \put(88.0, 1.4){\color{black}\scriptsize{{Clean image $\mathbf{x}_0$}}}
        \put(69.1, 17.4){\color{black}\scriptsize{{\textit{“sagittal lumber spine”}}}}
        \put(69.2, 19.8){\color{black}\scriptsize{{\textbf{Style-agnostic prompt}}}}
        % \put(69.5, 13.2){\color{black}\scriptsize{Ctrl}}
        % \put(72.2, 2){\color{black}\scriptsize{$U^{D_0}$}}
        \put(90.5, 26.){\color{black}\scriptsize{{Frozen}}}
        \put(90.5, 23.8){\color{black}\scriptsize{{Trainable}}}
        \put(90.5, 21){\color{black}\scriptsize{{$U^B$}}}
        \put(90.5, 19){\color{black}\scriptsize{{$U^{D_0}$}}}
        \put(90.5, 17){\color{black}\scriptsize{{ControlNet}}}
        \put(90.5, 15){\color{black}\scriptsize{{Seg. Model}}}
        \end{overpic}
        \caption{Training style intervention module.}
        \label{fig2_a}
    \end{subfigure}
    % \hfill
    \begin{subfigure}{0.95\linewidth}
        \centering
        \begin{overpic}[width=1 \columnwidth]{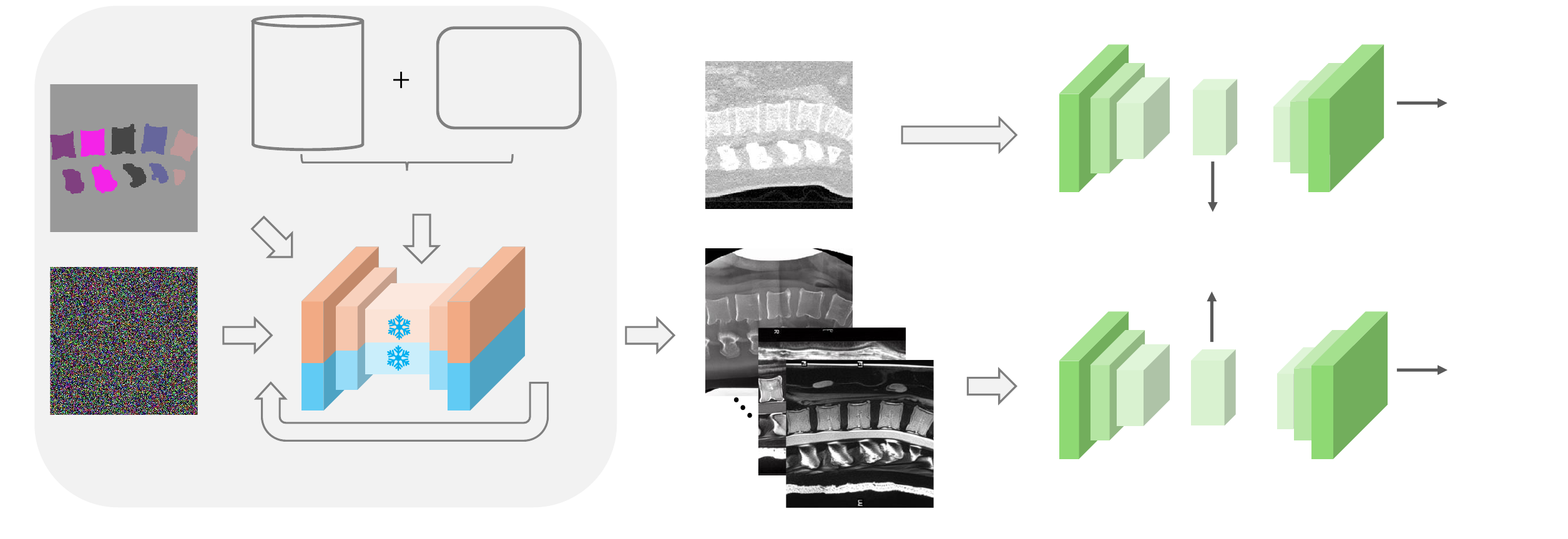}
        \put(16.7, 31){\color{black}\scriptsize{{\textit{“X-ray”}}}}
        \put(16.7, 29){\color{black}\scriptsize{{\textit{“T1MRI”}}}}
        \put(16.7, 27){\color{black}\scriptsize{{\textit{“T2MRI”}}}}
        \put(19., 26){\color{black}\scriptsize{{...}}}
        % \put(17.5, 21){\color{black}\scriptsize{{Style pool}}}
        \put(29.5, 30.5){\color{black}\scriptsize{{\textit{“sagittal}}}}
        \put(30.2, 29){\color{black}\scriptsize{\textit{{lumber}}}}
        \put(30.2, 27.5){\color{black}\scriptsize{{\textit{spine} \{\}\textit{”}}}}
        \put(18.3, 22.1){\color{black}\scriptsize{{\textbf{Style-intervention prompt}}}}
        % \put(19.5, 17.2){\color{black}\scriptsize{Ctrl}}
        % \put(24., 7){\color{black}\scriptsize{$U^{D_0}$}}
        \put(23.2, 8.2){\color{black}\scriptsize{$T$ steps}}
        % \put(76.2, 29){\color{black}\scriptsize{$U^{\text{Seg}}$}}
        % \put(76., 5){\color{black}\scriptsize{$U^{\text{Seg}}$}}
        \put(74.8, 18.2){\color{black}\footnotesize{$\mathcal{L}_{\text{InfoNCE}}$}}
        \put(93., 27.3){\color{black}\footnotesize{$\mathcal{L}_{\text{Seg}}$}}
        \put(93., 10.3){\color{black}\footnotesize{$\mathcal{L}_{\text{Seg}}$}}
        \put(5.1, 6.2){\color{black}\scriptsize{{Noise $\mathbf{x}_T$}}}
        \put(3, 18.1){\color{black}\scriptsize{{Conditional image $c_i$}}}
        \put(8.5, 3.8){\color{black}\footnotesize{{\textbf{Conditional diffusion augmentation}}}}
        \put(44.8, 19.6){\color{black}\scriptsize{{Source image $X$}}}
        \put(44.8, 0.1){\color{black}\scriptsize{{Augmented images $X^{do(S=S_i)}$}}}
        % \put(46, 0.1){\color{black}\scriptsize{{Augmented images $X^{do(S=S_i)}$}}}
        \end{overpic}
        \caption{Training segmentation model.}
        \label{fig2_b}
    \end{subfigure}
    \caption{Overview of our method. (a) Fine-tune pretrained SD U-Net ($U^{B}$) on the source domain $D_0$ (left), and then train ControlNet with the fine-tuned SD U-Net ($U^{D_0}$) to inject image conditions (right), both using style-agnostic prompts. (b) Generate style-intervened images using ControlNet and $U^{B}$ with style-intervention prompts. The segmentation model is trained on pairs of original and style-intervened images using a segmentation loss and InfoNCE regularization.}
    \label{fig2}
\vspace{-0.5cm}
\end{figure*}
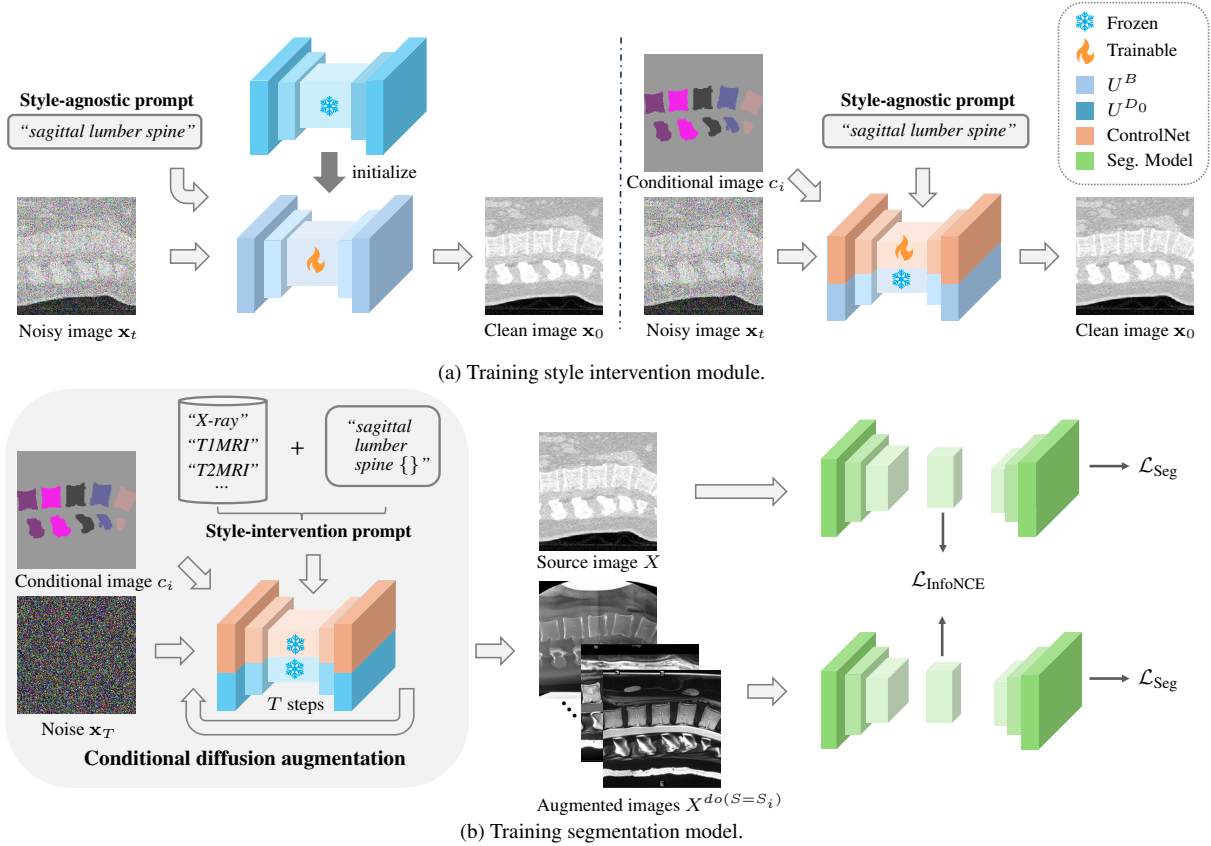

\subsection{Single-source Domain Generalization under a Causal Perspective}
Let $X \in \mathcal{X}$ and $Y \in \mathcal{Y}$ represent input images and corresponding segmentation masks. We extend common assumptions~\cite{lv2022causality, ouyang2022causality} on the data generative process from a causal perspective by allowing an additional causal relation from content to style as proposed in~\cite{von2021self} (\cref{fig1_a}). We further assume that an expert (or oracle) is able to provide correct segmentation masks $Y$ from observations $X$ alone. 

In the context of \gls{sdg}, we assume that we are given training pairs from a single domain $D_0$: $\{(X^0_i, Y^0_i)\}_{i=1}^{n_0}$, and a fixed set of target domains $D_1, \ldots, D_N$. Each dataset $D_e$ contains \gls{iid} samples from some probability distribution $P(X^e, Y^e)$. {We aim to obtain an optimal predictor $f$ to enable \gls{ood} generalization. In particular, $f$ is trained on $D_0$ such that $f$ minimizes the worst case risk $R^e(f) := \mathbb{E}_{X^e, Y^e} \left[ \ell(f(X^e), Y^e) \right]$ for any given target domain $D_e$.
Arjovsky et al.~\cite{arjovsky2019invariant} demonstrate that an invariant predictor would obtain an optimal solution if and only if it uses only the direct causal parents of $Y$ in the corresponding \gls{scm}~\cite{pearl2009causality}.}

% \begin{definition}[\cite{arjovsky2019invariant}]
% \label{def-1}
% \textnormal{A data representation \( \Phi : \mathcal{X} \rightarrow \mathcal{H} \) elicits an invariant predictor \( w \circ \Phi \) across environments (domains) \( \mathcal{E} \) if there is a classifier \( w : \mathcal{H} \rightarrow \mathcal{Y} \) simultaneously optimal for all environments, that is,
% \[
% w \in \arg \min_{\bar{w} : \mathcal{H} \rightarrow \mathcal{Y}} R^e(\bar{w} \circ \Phi) \quad \text{for all } e \in \mathcal{E}.
% \]}
% \end{definition}

% Arjovsky et al.~\cite{arjovsky2019invariant} propose an \gls{irm} algorithm to obtain this invariant solution from \textit{multiple} domains and show that this solution is optimal if and only if it uses only the direct causal parents of $Y$ in the corresponding \gls{scm}~\cite{pearl2009causality}. In this work, we demonstrate how to achieve this optimal solution given a \emph{single} source domain. 

Under the described SCM assumptions, von Kügelgen et al.~\cite{von2021self} prove that if the augmented pairs of views $(X, X^+)$ in contrastive methods are generated under the principle of a ``soft'' intervention on $S$, then the InfoNCE~\cite{oord2018representation} objective combined with an encoder function $\Phi$ (\eg, \gls{nn}) \textit{identifies} the invariant content $C$ partition: 
\begin{equation}\label{eq: InfoNCE}
    \mathcal{L}_{\text{InfoNCE}} = -\mathbb{E}_{X \sim p_X} \left[ \log \frac{\exp(\text{sim}(\nu, \nu^+) / \tau)}{\sum_{X^- \in \mathcal{N}} \exp(\text{sim}(\nu, \nu^-) / \tau)} \right],
\end{equation}
where $\nu = \Phi(X)$, $\text{sim}(\cdot,\cdot)$ is some similarity metric (\eg, cosine similarity), $\tau$ is a temperature parameter, and $X^-$ are negative pairs.

The notion of \textit{identifiability} is crucial here, as it means that under suitable augmentations, the learned representations are expected to match the true underlying latent content factors - the essential ingredient for the optimal solution of \gls{dg} problem~\cite{arjovsky2019invariant}. 

To adapt to the segmentation task and retain the spatial structure by patch-wise contrastive learning, we apply the InfoNCE loss after the bottleneck layer of the U-Net and combine it with segmentation loss (such as cross-entropy or Dice loss~\cite{drozdzal2016importance}): 

\begin{equation}\label{eq: combined_loss}
    \mathcal{L}_{\text{combined}} = \mathcal{L}_{\text{Seg}} + \lambda_{\text{reg}} \mathcal{L}_{\text{InfoNCE}},
\end{equation}
{where \(\lambda_{\text{reg}}\) is a hyperparameter that balances the contribution of the InfoNCE loss.}

In real-world applications, direct intervention on the style variable is infeasible as it is unobserved. For instance, in medical imaging, this would require scanning the same patient using different imaging modalities multiple times under controlled conditions—a process that is often impractical for acquiring training data. However, Ilse et al.~\cite{ilse2021selecting} introduce the concept of \textit{intervention-augmentation equivariance}, formally demonstrating that data augmentation \textit{can} serve as a surrogate tool for simulating interventions.

In the following section, we apply this principle to design augmentation pairs in~\cref{eq: InfoNCE}, enabling the InfoNCE-regularized training to recover content features and obtain the optimal predictor described above. 
% We apply this principle to design augmentation pairs in~\cref{eq: InfoNCE}, such that InfoNCE-regularized training is able to recover content features and we can obtain an optimal predictor described above. 
% In the following section, we illustrate the design of an augmentation such that it satisfies~\cref{def-2}. Specifically, rather than conducting real interventions, we employ carefully controlled diffusion models to simulate these interventions, as if the same patient were measured multiple times with various imaging modalities.
% We thus apply this principle to design augmentations pairs in \cref{eq: InfoNCE}, such that InfoNCE-regularized training is able to recover content features and we can obtain an optimal predictor described above. We demonstrate in the following section how to design an augmentation such that it satisfies \cref{def-2}. More specifically, instead of performing a real intervention as described above, we use carefully controlled diffusion models to simulate such intervention as if the same patient has been measured multiple times using different imaging modalities.

\subsection{Diffusion-based Style Intervention}

\subsubsection{Diffusion-based Image Generation}
% Diffusion models are known for their generative ability, beating previous generative models including GANs and VAEs, thanks to their iterative generation process and stable training. 

Recent advancements in text-to-image \glspl{ldm}, such as \gls{sd}~\cite{rombach2022high}, show the ability to generate diverse and realistic images. The rich prior knowledge learned from vast text-image datasets is widely exploited for downstream tasks such as data augmentation~\cite{trabucco2023effective,fang2024data}. An \gls{ldm} comprises a \gls{vae}~\cite{kingma2013auto} which projects images into a latent space and a U-Net~\cite{ronneberger2015u} that learns the denoising process within this space. Specifically, the diffusion process gradually introduces random Gaussian noise to the latent representation $\mathbf{z}_0$ of a clean image $\mathbf{x}_0$ over $T$ discrete steps~\cite{sohl2015deep, song2019generative}. For any intermediate noisy state, we can express it formally using the property of Gaussian noise as~\cite{ho2020denoising}:
\begin{equation}\label{eq:ddpm_forward_arbitrary}
\mathbf{z}_{t} = \sqrt{\alpha_t} \mathbf{z}_{0} + \sqrt{1-\alpha_t} \mathbf{\epsilon},
\end{equation}
where $t\in \{1,\cdots, T\}$ represents the current timestep or noise level, $\epsilon$ is a Gaussian noise, and $\alpha_t$ is a predefined noise schedule~\cite{ho2020denoising,song2020denoising}. Each training step estimates the noise added to $\mathbf{z}_t$ at a randomly selected timestep $t$ by optimizing the following regression loss~\cite{ho2020denoising, rombach2022high}:
% \begin{equation}\label{DDPM_loss}
% \mathcal{L}_{\text{SD}} = \mathbb{E}_{\mathbf{z}_0,t,c_t,\epsilon}\left[\gamma_t||\epsilon - {\epsilon}_\theta(\sqrt{\alpha_t} \mathbf{z}_{0} + \sqrt{1-\alpha_t} \mathbf{\epsilon},c_t, t)||^2\right],
% \end{equation}
\begin{equation}\label{DDPM_loss}
\mathcal{L}_{\text{SD}} = \mathbb{E}_{\mathbf{z}_0,t,c_t,\epsilon}\left[\gamma_t||\epsilon - {\epsilon}_\theta(\mathbf{z}_{t},c_t, t)||_2^2\right],
\end{equation}
where $t$ is uniformly sampled from $[0, T]$. The coefficient $\gamma_t$ adjusts the weight for different noise levels, $\mathbf{\epsilon}_\theta$ is the \gls{nn} that predicts the added noise, and $c_t$ is the text prompt that describes the image. During inference, we iteratively denoise a randomly sampled Gaussian noise $\mathbf{z}_T \sim \mathcal{N}(0, I)$ to generate $\mathbf{z}_0$ based on the prompt $c_t$, which is then decoded into an image $\mathbf{x}_0$.
% where $t$ is uniformly sampled over $[0, T]$, $\gamma_t$ is a weight for different noise levels, $\mathbf{\epsilon}_\theta$ is the neural network that predicts the added noise, and $c_t$ is the text prompt condition that describes the image. During inference, we iteratively denoise a randomly-sampled Gaussian noise $\mathbf{z}_T \sim \mathcal{N}(0, I)$ to get a new $\mathbf{z}_0$ following prompt $c_t$, which is later decoded into an image $\mathbf{x}_0$. 

%% intro to diffusion sampling process
% The sampling process aims to generate a clean image from Gaussian noise $z_T \sim \mathcal{N}(\mathbf{0}, \mathbf{I})$, and each reverse step is defined by:
% \begin{equation}\label{eq:ddim_reverse}
% \begin{aligned}
% \mathbf{z}_{t-1}=&\sqrt{{\alpha}_{t-1}}\left(\frac{\mathbf{z}_{t}-\sqrt{1-{\alpha}_{t}} \mathbf{\epsilon}_\theta(\mathbf{z}_{t}, t)}{\sqrt{{\alpha}_{t}}}\right)\\
% +&\sqrt{1-{\alpha}_{t-1}-{\sigma_{{\eta}_t}}^{2}} \cdot \mathbf{\epsilon}_\theta(\mathbf{z}_{t}, t)+\sigma_{{\eta}_t} \mathbf{\epsilon}_{t},
% \end{aligned}
% \end{equation}
% where $\mathbf{\epsilon}_{t}$ is standard Gaussian noise, inside the parentheses of the first term is the predicted $\mathbf{x}_{0}$ at timestep $t$, and the magnitude of $\sigma_{{\eta}_t}$ controls how stochastic the forward process is.

\subsubsection{Controllable Generation for Style Intervention}
{To achieve intervention-augmentation equivariance, we assume the segmentation mask represents the content variable in the causal data generation process describe in~\cref{fig1_a}. Since we wish to intervene only on the style variable, the augmentations should share the same segmentation masks as source domain images. Such a constraint is difficult to satisfy via only prompting on the anatomy. To this end, we add a ControlNet module~\cite{zhang2023adding}. The training objective of this module is to teach the model to denoise data given a conditional image (\eg, semantic segmentation mask, etc.), which ensures that the model learns to preserve condition-specific contents during the generation process. In our case, we train the ControlNet module with source domain images and corresponding segmentation masks such that it captures the relationship between generated anatomical structures and given segmentation masks. Then, we can leverage the generative prior of the pretrained \gls{sd} and the content-preserving capabilities of the ControlNet to generate style-varied images while maintaining the segmentation-relevant content.}
% While \gls{sd} allows for the generation of images in various styles and anatomies through prompting,nit does not preserve image content. Thus, to intervene only on the style variable,  we incorporate the segmentation masks as image conditions to obtain pixel-aligned images.
% into the image generation process to obtain 
% images that are pixel-aligned with the segmentation masks.
% pixel-aligned images.
% The stable diffusion model can be fine-tuned to do style transfer by telling the model the prompt instruct \cite{brooks2023instructpix2pix}, or by specifying the content that wants to be preserved \cite{zhang2023adding, mou2024t2i}; besides the text description, the target style can also be enhanced by providing a reference image with the desired style \cite{zhang2023inversion}.
% For this purpose, we utilize the ControlNet module \cite{zhang2023adding}, which enables the integration of image conditions through an additional trainable module. 
%This module's outputs feed into a fixed \gls{sd} backbone, ensuring content preservation.\Yuanzhi{modify here} 
% This module will provide additional control signal, \ie, the segmentation masks to the \gls{sd} backbone $U^B$, ensuring content preservation.

Specifically, the ControlNet module is initialized as a copy of the encoding layers of the \gls{sd} U-Net and connected via zero convolutions.
% \Boqi{@yuanzhi: can you modify here. Sometimes it's ControlNet framework and sometimes it's ControlNet module. Can you make it consistent?}
% By training the ControlNet with the 
% As a result, we can only intervene on style variables by prompting different styles to make use of 2d prior stored in the pretrained SD model.
The training of the ControlNet follows the loss defined in \cref{DDPM_loss} with an additional segmentation mask condition $c_i$:
\begin{equation}\label{eq: controlnet_training}
\mathcal{L}_{\text{ControlNet}} = \mathbb{E}_{\mathbf{z}_0, t, c_t, c_i, \epsilon} \left[ \left\| \epsilon - \epsilon_\theta(\mathbf{z}_t, t, c_t, c_i) \right\|_2^2 \right],
\end{equation}
here $\theta$ represents the parameters of both the trainable ControlNet module and the frozen pretrained \gls{sd} backbone U-Net ($U^B$). 
After training, the ControlNet module will provide additional control signal, \ie, the segmentation masks, to $U^B$, ensuring content preservation in the generation process.
% \Boqi{A key advantage here is we only condition on the foreground, \ie, segmentation masks, allowing variate of all background pixels. Compared to style-transfer-based methods, such as AdaIn which has no control over the disentanglement of foreground and background, all changes together, introduce spurious correlation}

% \Yuanzhi{polish}
Jia et al.~\cite{jia2023dginstyle} empirically show that ControlNet, when trained with $U^B$ on the source domain $D_0$ directly, learns to inject not only image content conditions but also style information of $D_0$. 
% learns to inject not only image content condition but also style information of the source domain $D_0$ when training the ControlNet with pre-trained base \gls{sd} U-Net $U^B$ directly.
We conjure this undesired style overfitting is due to a mismatch between the prior information in $U^B$ and the domain knowledge of $D_0$ used to train the ControlNet, which compromises the style prior stored in $U^B$. To mitigate this issue, we adopt the Style Swap technique~\cite{jia2023dginstyle}, forcing ControlNet to focus on learning the content information. Specifically, instead of training ControlNet directly on $D_0$ with $U^B$, we first fine-tune $U^B$ on $D_0$ using instance-fine-tuning strategies~\cite{gal2022image, ruiz2023dreambooth, han2023svdiff, wang2024instantid} to obtain an instance fine-tuned U-Net ($U^{D_0}$). Following~\cite{jia2023dginstyle}, we employ DreamBooth~\cite{ruiz2023dreambooth} which adjusts \gls{sd} U-Net parameters to recognize specific instances by optimizing the following objective:
% We specifically use Dreambooth~\cite{ruiz2023dreambooth} following~\cite{jia2023dginstyle}, which adjusts \gls{sd} U-Net parameters to recognize specific instances:
\begin{equation}\label{eq:db_training}
\begin{split}
    \mathcal{L}_{U^{D_0}} =  \mathbb{E}_{\mathbf{z_0}, t, {c_t}, \epsilon, \epsilon'} \Bigg[ \gamma_t \left\| \hat{\mathbf{z}}_\theta (\mathbf{z}_t , {c_t}, t) - \mathbf{z}_0 \right\|_2^2 \\
    + \lambda_{\text{pr}} \gamma_{t'} \left\| \hat{\mathbf{z}}_\theta (\mathbf{z}^{pr}_{t'} , {c_t}^{\text{pr}}, t') - \mathbf{z}^{\text{pr}}_0 \right\|_2^2 \Bigg],
\end{split}
\end{equation}
where $\hat{\mathbf{z}}_\theta$ is the clean signal prediction network \cite{rombach2022high} (which is equivalent to noise prediction $\epsilon_\theta$ up to an affine transformation), $\mathbf{z_0} \in D_0$ is the data of new instance, and $\mathbf{z}^{\text{pr}}_0$ is the generated samples of the pre-trained model using text condition ${c_t}^{\text{pr}}$. The second term is the prior-preservation term and $\lambda_{\text{pr}}$ controls the relative weight of this term.

\subsection{Overview of Methods}
\cref{fig2} illustrates the overview of our method. It consists of two stages: training the style intervention module and the segmentation model. In the first stage (\cref{fig2_a}), we fine-tune a pretrained \gls{sd} U-Net ($U^{B}$) on the source domain $D_0$ using style-agnostic prompts. These prompts guide image generation with information on anatomy but not imaging modality, resulting in an instance-tuned U-Net ($U^{D_0}$) that overfits to the style of $D_0$. We then train the ControlNet alongside $U^{D_0}$ on $D_0$, focusing on learning to inject image conditions into the generation process. In the second stage (\cref{fig2_b}), we integrate the original $U^{B}$ with ControlNet to leverage its rich style prior and use segmentation labels from $D_0$ as image conditions. To construct the style-intervention prompt, we concatenate unique class names of the anatomies present in the segmentation masks with a style prompt randomly sampled from a predefined style pool. This prompt serves as the text condition. This approach enables us to generate training pairs of original and style-intervened images. 
Finally, these pairs are used to train the segmentation model, which is optimized using the loss function defined in~\cref{eq: combined_loss}.

\section{Experiments}
\subsection{Experimental Setting}
\label{exp_setting}
\noindent\textbf{Datasets and Preprocessing.} We evaluate our method on three tasks: \gls{as}, \gls{lss} and \gls{ls} with three imaging modalities: CT, MRI and X-Ray. For the \gls{as} task, we segment four abdominal organs: liver, right kidney (R-Kidney), left kidney (L-Kidney), and spleen. We use the CT and MRI data from~\cite{landman2015miccai} and~\cite{ji2022amos}, following the split in~\cite{ouyang2022causality} and~\cite{ji2022amos}, respectively. For the \gls{lss} task, we perform instance segmentation of the vertebrae from L1 to L5. We construct a cross-modality dataset using CT scans from~\cite{sekuboyina2020labeling}, T2-MRI from~\cite{pang2020spineparsenet} with 30 volumes of in-house T1/T2 MRI scans, and X-Ray scans from~\cite{klinwichit2023buu}. For the \gls{ls} task, we segment left lung (L-Lung) and right lung (R-Lung) with CT data from~\cite{yang2017data} and X-Ray data from~\cite{danilov2022automatic}. All 3D volumes are reformatted to 2D and resized to $256\times256$ pixels during training. {During inference, we obtain a 3D volume by stacking 2D slice predictions and compare it with the ground-truth 3D volume to calculate the per-volume Dice score. The final result is the average across all test volumes.} Common augmentations proposed in~\cite{zhang2020generalizing} is applied for all methods. 

\vspace{0.2cm}
\noindent\textbf{Implementation Details.} We use U-Net~\cite{ronneberger2015u} as our segmentation model and Dice loss~\cite{drozdzal2016importance} as the segmentation loss. For all datasets, we train the model for 50 epochs with a batch size of 64. We utilize the Adam optimizer with an initial learning rate of $1\times10^{-3}$. To ensure stable training, we decay the learning rate following the cosine annealing schedule. For the style intervention module, we fine-tune the \gls{sd} model trained on the ROCO dataset~\cite{pelka2018radiology} using DreamBooth for 15,000 iterations with a learning rate of $1\times10^{-6}$ and batch size of 1, and train the ControlNet for 72,000 iterations with a learning rate of $1\times10^{-5}$ and batch size of 2. All slices are resized to $512\times512$ pixels to meet the training resolution requirements of \gls{sd}. All experiments are conducted using PyTorch framework on a single NVIDIA RTX A6000 GPU with 48 GB of memory. Further details are provided in the supplemental material.
% During the sampling process, we adopt the UniPC sampler to accelerate the generation with 20 steps\cite{zhao2024unipc}. 
\begin{table*}[!t]
\footnotesize
\centering
\caption{Comparison of methods on the cross-modality abdominal segmentation (AS) task. Results are mean ± standard deviation over 3 random initializations. Best and second-best per column (excluding Baseline and Benchmark) are bold and underlined.}
{
\setlength{\tabcolsep}{0.65mm}{
\begin{tabular}{c|ccccc|ccccc}
    \toprule[0.8pt]
    \multirow{2}{*}{Method} & \multicolumn{5}{c|}{Abdominal CT $\rightarrow$ Abdominal MRI} & \multicolumn{5}{c}{Abdominal MRI $\rightarrow$ Abdominal CT}\\
    \cmidrule{2-11}
    & Liver & R-Kidney & L-Kidney & Spleen & Average & Liver & R-Kidney & L-Kidney & Spleen & Average \\
    \midrule[0.5pt]
    \textsc{Baseline} & 52.57\std{2.8} & 32.53\std{6.5} & 57.23\std{1.3} & 37.07\std{4.9} & 44.87\std{2.9} & 91.27\std{0.6} & 67.77\std{1.9} & 77.70\std{2.0} & 88.13\std{1.8} & 81.20\std{0.4} \\
    \midrule[0.5pt]
    \textsc{MixStyle}~\cite{zhou2021mixstyle} & 89.83\std{1.9} & 85.97\std{3.4} & 81.30\std{7.9} & 67.73\std{8.7} & 81.23\std{1.7} & 87.03\std{1.7} & 81.03\std{2.2} & 72.60\std{4.9} & 86.23\std{1.3} & 81.73\std{1.1} \\
    \textsc{RandConv}~\cite{xu2020robust} & 82.50\std{2.9} & 78.37\std{0.2} & 82.27\std{2.9} & 72.00\std{2.8} & 78.77\std{1.0} & 90.13\std{1.7} & 84.10\std{0.7} & 85.60\std{1.9} & 87.57\std{0.9} & 86.87\std{0.8} \\
    \textsc{DualNorm}~\cite{zhou2022dn} & 82.90\std{1.7} & 80.17\std{3.1} & 73.33\std{0.8} & 55.10\std{2.1} & 72.87\std{1.0} & 84.47\std{1.3} & 80.80\std{6.6} & 79.97\std{6.7} & 80.17\std{5.7} & 81.60\std{3.5} \\
    \textsc{CSDG}~\cite{ouyang2022causality} & \underline{87.27\std{0.7}} & 84.93\std{0.0} & 87.00\std{0.9} & 73.67\std{1.9} & 83.20\std{0.7} & 91.77\std{0.1} & \textbf{86.63\std{0.7}} & 89.00\std{0.4} & \underline{88.93\std{0.6}} & 89.07\std{0.1} \\
    \textsc{SLAug}~\cite{su2023rethinking} & 86.73\std{0.8} & \underline{86.60\std{2.2}} & \textbf{89.97\std{0.3}} & \underline{79.27\std{0.8}} & \underline{85.67\std{0.9}} & \underline{92.97\std{0.4}} & 86.07\std{0.9} & \textbf{90.30\std{0.7}} & 87.50\std{0.7} & \underline{89.17\std{0.5}} \\
    \textsc{Ours} & \textbf{89.57\std{1.2}} & \textbf{87.60\std{0.9}} & \underline{87.30\std{0.6}} & \textbf{80.33\std{0.6}} & \textbf{86.20\std{0.3}} & \textbf{92.97\std{0.2}} & \underline{86.27\std{1.1}} & \underline{90.20\std{1.1}} & \textbf{91.33\std{0.7}} & \textbf{90.17\std{0.7}} \\
    \midrule[0.5pt]
    \textsc{Benchmark} & 96.27\std{0.2}& 95.83\std{0.2} & 95.50\std{0.2} & 94.97\std{0.2} & 95.63\std{0.1} & 95.50\std{0.2} & 85.20\std{1.1} & 91.20\std{1.2} & 93.17\std{0.1} & 91.30\std{0.5} \\
    \bottomrule[0.8pt]
\end{tabular}
}
}
\label{tab:1}
\vspace{-0.2cm}
\end{table*}

\subsection{Results and Discussion}
We evaluate our method against multiple baselines: a weak baseline consisting of a segmentation model trained on the source domain with only basic augmentations; a supervised benchmark where the segmentation model is trained on the target domain; and several \gls{dg} methods. In particular, MixStyle~\cite{zhou2021mixstyle} enhances generalization by mixing styles of training instances, while RandConv~\cite{xu2020robust} improves robustness via randomly initialized convolution layers. We also compare with three \gls{sdg} methods for medical image segmentation. DualNorm~\cite{zhou2022dn} employs separate normalization layers for domain-similar and domain-dissimilar augmentations generated via nonlinear Bézier transformations~\cite{zhou2019models}. CSDG~\cite{ouyang2022causality} uses a shallow network to generate novel training samples, mixes them via pseudo-correlation maps, and learns domain-invariant features using a \gls{kl} loss-based regularizer to remove spurious correlations. SLAug~\cite{su2023rethinking} enforces class-level representation invariance using constrained Bézier transformations on both global and class-level regions, fusing the locally and globally transformed augmentations using a saliency-balancing approach. Implementation details are provided in the supplementary material. Preprocessing and augmentation are consistent across all methods, as described in~\cref{exp_setting}.

\subsubsection{Comparison with State-of-the-art Methods}
The quantitative results for cross-modality \gls{as}, \gls{lss} and \gls{ls} are summarized in~\cref{tab:1},~\cref{tab:2}, and~\cref{tab:3}, respectively. Overall, our approach {consistently} outperforms state-of-the-art methods across all tasks. Specifically, in the \gls{as} task, our method achieves the highest average Dice scores of 86.20 \gls{pp} and 90.17 \gls{pp} in the CT to MRI and MRI to CT directions, respectively. {Although the improvement over the most competitive baseline, SLAug, is modest (0.53 \gls{pp} and 1.00 \gls{pp}), our approach consistently approaches the supervised benchmark, with only 1.13 \gls{pp} difference for source domain MRI.} In the \gls{lss} task, our method surpasses the previous best-performing methods by 3.13 \gls{pp} (CT to MRI) and 4.03 \gls{pp} (MRI to CT). {The more substantial improvements compared with the \gls{as} task showcase our method’s superiority in generalizing across large domain gaps. In the \gls{ls} task, our method demonstrates an improvement of over 10 \gls{pp} in Dice score compared to the baseline, whereas other methods show only marginal (less than 1 \gls{pp}) or even negative improvements. Specifically, we achieve an average Dice score of 78.79 \gls{pp}, which is 10.78 \gls{pp} higher than the second-best method (SLAug, 68.01 \gls{pp}). This substantial gain highlights the robustness of our approach, particularly in scenarios with complex domain shifts where other \gls{dg} techniques struggle. These consistent performance gains across various tasks and modalities demonstrate that our approach effectively handles both small and large domain gaps, offering superior generalizability across diverse anatomical structures and imaging modalities in medical image segmentation.}

\begin{figure}[t]
\centering
\begin{overpic}[width=1. \columnwidth]{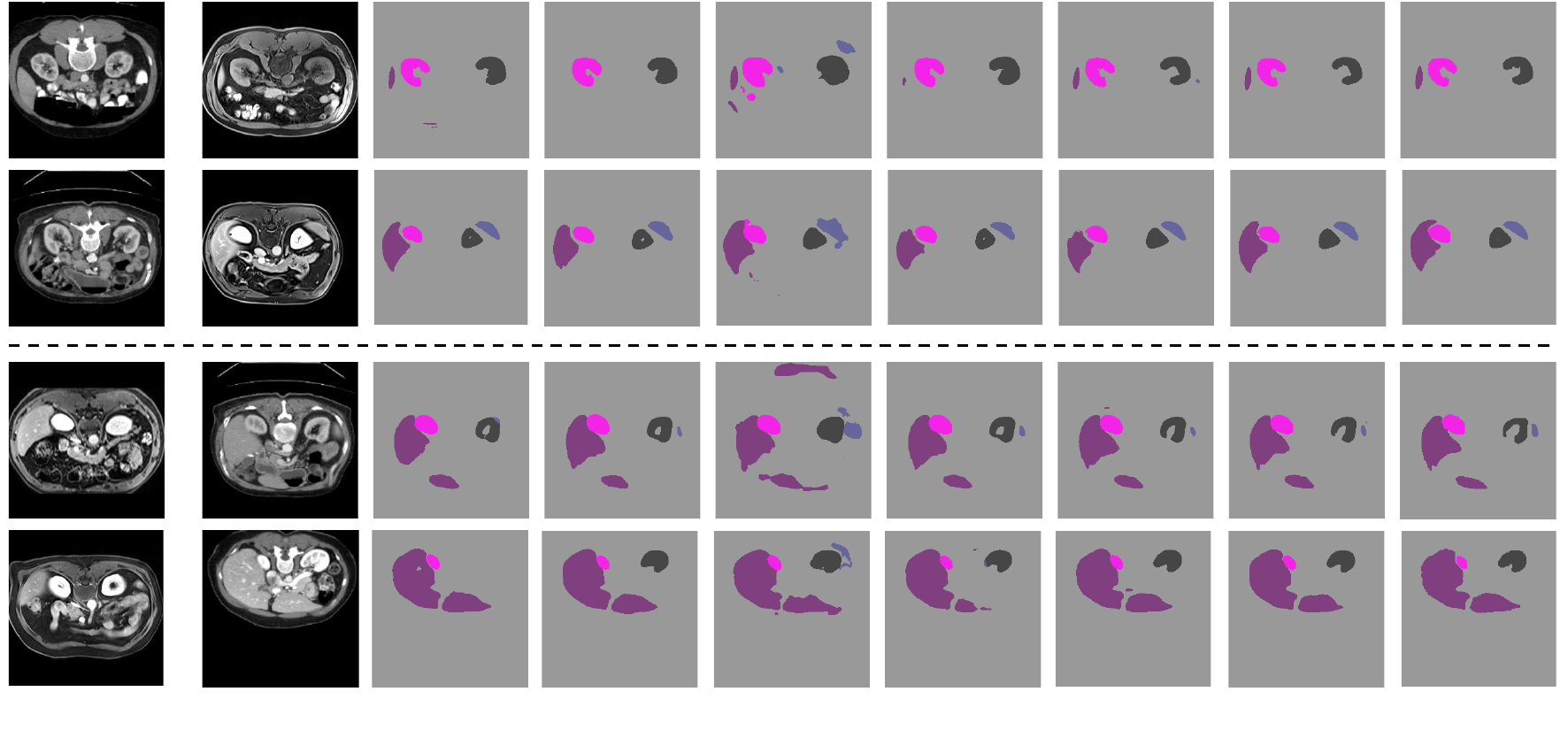}
    \put(2.8, 0.5){\color{black}\tiny{Source}}
    % \put(3.0, -2){\color{black}\tiny{Image}}
    \put(15.25, 0.5){\color{black}\tiny{Target}}
    % \put(13.75, -2.){\color{black}\tiny{Image}}
    \put(25.0, 0.5){\color{black}\tiny{MixStyle}}
    \put(35.1, 0.5){\color{black}\tiny{RandConv}}
    \put(46.1, 0.5){\color{black}\tiny{DualNorm}}
    \put(58.5, 0.5){\color{black}\tiny{CSDG}}
    \put(69.25, 0.5){\color{black}\tiny{SLAug}}
    \put(81.0, 0.5){\color{black}\tiny{Ours}}
    \put(89.0, 0.5){\color{black}\tiny{Ground Truth}}
\end{overpic}
\caption{Visualization of segmentation results on the \gls{as} task.. First two rows: ``CT to MRI'' task; Last two rows: ``MRI to CT''.}
\label{fig3}
\vspace{-10pt}
\end{figure}

\begin{figure}[t]
\centering
\begin{overpic}[width=1. \columnwidth]{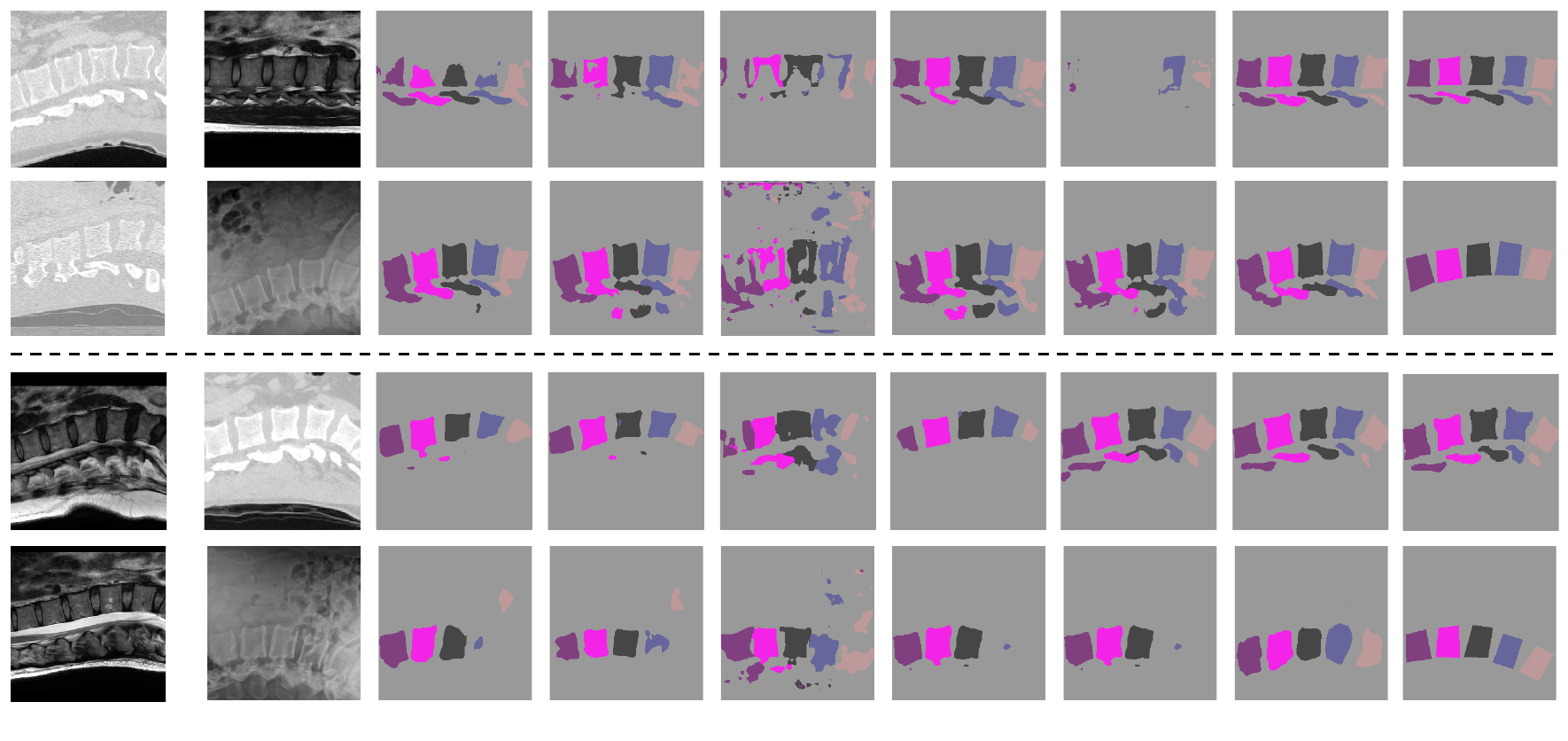}
    \put(2.8, 0.5){\color{black}\tiny{Source}}
    % \put(3.0, -2){\color{black}\tiny{Image}}
    \put(15.25, 0.5){\color{black}\tiny{Target}}
    % \put(13.75, -2.){\color{black}\tiny{Image}}
    \put(25.0, 0.5){\color{black}\tiny{MixStyle}}
    \put(35.1, 0.5){\color{black}\tiny{RandConv}}
    \put(46.1, 0.5){\color{black}\tiny{DualNorm}}
    \put(58.5, 0.5){\color{black}\tiny{CSDG}}
    \put(69.25, 0.5){\color{black}\tiny{SLAug}}
    \put(81.0, 0.5){\color{black}\tiny{Ours}}
    \put(89.0, 0.5){\color{black}\tiny{Ground Truth}}
\end{overpic}
\caption{Visualization of segmentation results on the \gls{lss} task. First row: ``CT to MRI''; second row: ``CT to X-Ray''; third row: ``MRI to CT''; and last row: ``MRI to X-Ray''.}
\label{fig4}
\vspace{-10pt}
\end{figure}

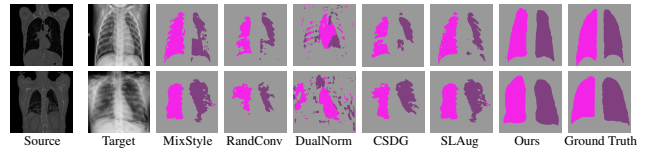
\begin{figure}[t]
\centering
\begin{overpic}[width=1. \columnwidth]{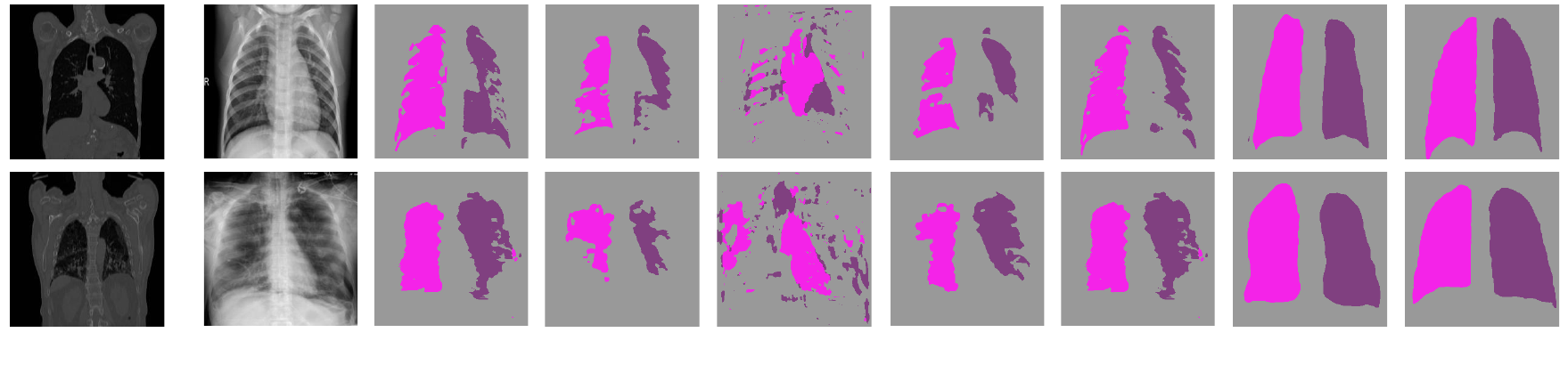}
    \put(2.8, 0.5){\color{black}\tiny{Source}}
    % \put(3.0, -2){\color{black}\tiny{Image}}
    \put(15.25, 0.5){\color{black}\tiny{Target}}
    % \put(13.75, -2.){\color{black}\tiny{Image}}
    \put(25.0, 0.5){\color{black}\tiny{MixStyle}}
    \put(35.1, 0.5){\color{black}\tiny{RandConv}}
    \put(46.1, 0.5){\color{black}\tiny{DualNorm}}
    \put(58.5, 0.5){\color{black}\tiny{CSDG}}
    \put(69.25, 0.5){\color{black}\tiny{SLAug}}
    \put(81.0, 0.5){\color{black}\tiny{Ours}}
    \put(89.0, 0.5){\color{black}\tiny{Ground Truth}}
\end{overpic}
\caption{Visualization of segmentation results on the \gls{ls} task. First two rows: ``CT to X-Ray''.}
\label{fig5}
\vspace{-10pt}
\end{figure}

We also visualize the segmentation results of our method and other \gls{dg} approaches for all tasks in~\cref{fig3},~\cref{fig4} and~\cref{fig5}, respectively. The first two columns show source and target domain images, illustrating the domain shift. We can see that the segmentation masks in the unseen target domain produced by our method exhibit greater accuracy and improved spatial continuity of the foreground compared to the baselines.

\begin{table*}[!t]
\footnotesize
\centering
\caption{Comparison of methods on the cross-modality lumbar spine segmentation (LSS) task. Results are mean ± standard deviation over 3 random initializations. Best and second-best per column (excluding Baseline and Benchmark) are bold and underlined.}
{
\setlength{\tabcolsep}{0.65mm}{
\begin{tabular}{c|cccccc|cccccc}
    \toprule[0.8pt]
    \multirow{2}{*}{Method} & \multicolumn{6}{c|}{Lumbar Spine CT $\rightarrow$ Lumbar Spine MRI} & \multicolumn{6}{c}{Lumbar Spine MRI $\rightarrow$ Lumbar Spine CT}\\
    \cmidrule{2-13}
    & L1 & L2 & L3 & L4 & L5 & Avg & L1 & L2 & L3 & L4 & L5 & Avg \\
    \midrule[0.5pt]
    \textsc{Baseline} & 0.00\std{0.0} & 0.00\std{0.0} & 0.00\std{0.0} & 0.00\std{0.0} & 0.25\std{0.0} & 0.05\std{0.4} & 0.05\std{0.1} & 0.60\std{0.7} & 1.57\std{1.4} & 1.63\std{1.8} & 0.00\std{0.0} & 0.80\std{0.7} \\
    \midrule[0.5pt]
    \textsc{MixStyle}~\cite{zhou2021mixstyle} & 41.90\std{6.6} & 37.20\std{2.0} & 39.60\std{7.9} & 48.57\std{6.7} & 47.33\std{6.7} & 42.93\std{2.6} & 42.83\std{3.7} & 48.77\std{5.4} & 50.43\std{7.5} & 43.63\std{1.7} & 36.87\std{2.9} & 44.47\std{3.9} \\
    \textsc{RandConv}~\cite{xu2020robust} & 65.93\std{0.8} & 66.07\std{0.6} & 65.00\std{1.2} & 66.63\std{1.5} & 70.27\std{1.6} & 66.80\std{1.0} & 52.17\std{1.5} & 51.43\std{0.6} & 51.67\std{0.6} & 41.23\std{1.1} & 36.17\std{2.7} & 46.53\std{1.0} \\
    \textsc{DualNorm}~\cite{zhou2022dn} & 40.00\std{4.6} & 46.20\std{1.3} & 45.63\std{2.2} & 42.70\std{4.0} & 46.77\std{3.0} & 44.27\std{2.8} & 28.73\std{0.7} & 30.37\std{0.5} & 29.63\std{3.8} & 27.33\std{3.4} & 22.30\std{2.4} & 27.70\std{1.8} \\
    \textsc{CSDG}~\cite{ouyang2022causality} & \underline{69.70\std{2.1}} & \underline{70.47\std{1.1}} & \underline{70.23\std{0.8}} & \underline{73.10\std{0.7}} & \underline{75.53\std{0.4}} & \underline{71.80\std{0.9}} & 46.10\std{3.9} & 52.70\std{1.0} & 54.70\std{1.7} & 50.23\std{3.0} & 39.20\std{0.3} & 48.57\std{0.3} \\
    \textsc{SLAug}~\cite{su2023rethinking} & 43.80\std{5.5} & 36.90\std{3.7} & 35.60\std{3.0} & 44.10\std{8.2} & 44.40\std{11.6} & 40.97\std{6.0} & \underline{73.90\std{1.6}} & \underline{72.73\std{2.3}} & \underline{75.77\std{2.1}} & \underline{69.57\std{2.1}} & \textbf{54.43\std{0.9}} & \underline{69.27\std{1.8}} \\
    \textsc{Ours} & \textbf{72.23\std{1.6}} & \textbf{73.00\std{2.4}} & \textbf{74.20\std{1.2}} & \textbf{76.57\std{0.6}} & \textbf{78.70\std{0.4}} & \textbf{74.93\std{1.2}} & \textbf{80.50\std{0.4}} & \textbf{81.20\std{0.4}} & \textbf{77.70\std{0.6}} & \textbf{73.70\std{1.1}} & \underline{53.60\std{0.7}} & \textbf{73.30\std{0.1}} \\
    \midrule[0.5pt]
    \textsc{Benchmark} & 89.33\std{0.8} & 89.97\std{0.2} & 88.43\std{0.1} & 89.13\std{0.8} & 80.33\std{13.8} & 87.43\std{2.4} & 90.70\std{1.8} & 89.77\std{2.8} & 88.23\std{1.8} & 88.80\std{1.3} & 72.10\std{14.4} & 85.90\std{2.8} \\
    \bottomrule[0.8pt]
    \toprule[0.8pt]
     
    \multirow{2}{*}{Method} & \multicolumn{6}{c|}{Lumbar Spine CT $\rightarrow$ Lumbar Spine X-Ray} & \multicolumn{6}{c}{Lumbar Spine MRI $\rightarrow$ Lumbar Spine X-Ray}\\
    \cmidrule{2-13}
     
    & L1 & L2 & L3 & L4 & L5 & Avg & L1 & L2 & L3 & L4 & L5 & Avg \\
    \midrule[0.5pt]
     
    \textsc{Baseline} & 3.27\std{1.6} & 8.32\std{4.4} & 11.08\std{4.3} & 12.17\std{4.6} & 13.45\std{8.4} & 9.66\std{3.4} & 0.03\std{0.06} & 1.50\std{1.7}& 1.47\std{1.3} & 5.00\std{4.0} & 2.17\std{0.6} & 2.03\std{1.0} \\
    \midrule[0.5pt]
     
    \textsc{MixStyle}~\cite{zhou2021mixstyle} & 35.87\std{14.6} & 50.20\std{9.0} & 52.37\std{4.8} & 47.53\std{5.9} & 42.03\std{7.3} & 45.60\std{7.1} & 24.33\std{7.3} & 42.63\std{11.8} & 46.47\std{5.9} & 45.20\std{5.4} & 44.57\std{13.3} & 40.64\std{8.4} \\
     
    \textsc{RandConv}~\cite{xu2020robust} & \underline{67.17\std{1.5}} & \underline{72.53\std{1.3}} & \underline{72.03\std{1.2}} & \underline{71.43\std{1.4}} & \underline{72.13\std{0.8}} & \underline{71.06\std{0.9}} & \underline{49.83\std{2.8}} & 64.40\std{3.0} & 65.87\std{1.3} & 72.57\std{0.1} & 63.23\std{1.2} & 63.18\std{1.6} \\
     
    \textsc{DualNorm}~\cite{zhou2022dn} & 19.74\std{4.2} & 23.25\std{2.9} & 25.99\std{3.1} & 23.24\std{4.8} & 31.29\std{2.7} & 24.70\std{1.4} & 45.34\std{2.3} & 53.61\std{2.1} & 54.94\std{2.1} & 55.95\std{1.9} & 48.56\std{2.7} & 51.68\std{0.2} \\
     
    \textsc{CSDG}~\cite{ouyang2022causality} & 66.20\std{0.7} & 70.50\std{0.4} & 70.67\std{0.6} & 71.17\std{0.7} & \textbf{73.40\std{2.0}} & 70.39\std{0.5} & 49.20\std{0.4} & \underline{65.00\std{1.0}} & \underline{68.47\std{1.5}} & \textbf{73.80\std{0.4}} & \underline{65.53\std{0.9}} & \underline{64.40\std{0.1}} \\
     
    \textsc{SLAug}~\cite{su2023rethinking} & 62.80\std{4.0} & 69.13\std{2.2} & 67.70\std{0.6} & 67.80\std{0.6} & 72.10\std{0.7} & 67.91\std{1.3} & 42.87\std{3.0} & 48.50\std{5.8} & 51.73\std{5.6} & 55.67\std{7.8} & 55.33\std{5.7} & 50.82\std{5.1} \\
     
    \textsc{Ours} & \textbf{69.97\std{2.1}} & \textbf{75.23\std{0.9}} & \textbf{73.97\std{0.1}} & \textbf{74.63\std{0.7}} & 71.93\std{0.3} & \textbf{73.15\std{0.3}} & \textbf{63.40\std{1.6}} & \textbf{72.95\std{2.1}} & \textbf{75.00\std{1.1}} & \underline{73.75\std{3.0}} & \textbf{67.65\std{0.6}} & \textbf{70.55\std{0.2}} \\
    \midrule[0.5pt]
     
    \textsc{Benchmark} & 93.1\std{0.4} & 93.9\std{0.3} & 94.3\std{0.3} & 94.3\std{0.2} & 92.3\std{0.3}& 93.6\std{0.3} & 93.1\std{0.4} & 93.9\std{0.3} & 94.3\std{0.3} & 94.3\std{0.2} & 92.3\std{0.3}& 93.6\std{0.3} \\
    \bottomrule[0.8pt]
\end{tabular}
}
}
\label{tab:2}
\vspace{-0.3cm}
\end{table*}

\begin{table}[ht]
\centering
\footnotesize
\caption{Comparison of methods on the cross-modality lung segmentation (LS) task. Results are mean ± standard deviation over 3 random initializations. Best and second-best per column (excluding Baseline and Benchmark) are bold and underlined.}
\begin{tabular}{l|ccc}
    \toprule[0.8pt]
    \multirow{2}{*}{Method} & \multicolumn{3}{c}{Lung CT $\rightarrow$ Lung X-Ray} \\
    \cmidrule{2-4}
    & L-Lung & R-Lung & Average \\
    \midrule[0.5pt]
    \textsc{Baseline} & 59.05\std{0.6} & 75.60\std{0.1} & 67.33\std{0.3} \\
    \midrule[0.5pt]
    \textsc{MixStyle}~\cite{zhou2021mixstyle} & 55.63\std{7.1} & 75.83\std{1.5} & 65.73\std{4.2} \\
    \textsc{RandConv}~\cite{xu2020robust} & 50.03\std{0.3} & 71.30\std{3.8} & 60.67\std{2.0}  \\
    \textsc{DualNorm}~\cite{zhou2022dn} & 36.82\std{0.3} & 19.24\std{0.7} & 28.03\std{2.4}  \\
    \textsc{CSDG}~\cite{ouyang2022causality} & 52.14\std{1.6} & 66.71\std{9.1} & 59.43\std{3.7} \\
    \textsc{SLAug}~\cite{su2023rethinking} & \underline{57.06\std{0.2}} & \textbf{78.97\std{0.2}} & \underline{68.01\std{0.0}}\\
    \textsc{Ours} & \textbf{79.84\std{0.5}} & \underline{77.75\std{0.5}} & \textbf{78.79\std{0.3}}  \\
    \midrule[0.5pt]
    \textsc{Benchmark} & 94.46\std{0.8} & 96.32\std{0.7} & 96.39\std{0.8}  \\
    \bottomrule[0.8pt]
\end{tabular}
\label{tab:3}
\vspace{-0.3cm}
\end{table}

\subsubsection{Ablation Study}
\noindent\textbf{Efficacy of InfoNCE-based Regularization.}
We evaluate the effectiveness of using patch-wise InfoNCE loss for regularization through ablation studies with three variants: (1) training the segmentation model on original and style-intervened images without regularization (w/o Reg.); (2) applying \gls{kl} loss to the output logits from original and style-intervened images following~\cite{ouyang2022causality} (w/ \gls{kl}); and (3) using standard InfoNCE loss~\cite{chen2020simple} on average-pooled features (w/ S. InforNCE). Results in~\cref{tab:3} show training on augmented data without regularization results in significantly lower performance, with a Dice score 7.46 \gls{pp} below ours, highlighting the importance of domain-invariance regularization. Our approach outperforms \gls{kl}-based regularization by 3.63 \gls{pp}, demonstrating InfoNCE loss's superiority in extracting domain-invariant content features. The 4.56 \gls{pp} improvement over non-patch-wise InfoNCE loss suggests the value of preserving spatial structure for segmentation tasks within contrastive learning frameworks.

\begin{table}[ht]
\scriptsize
\centering
\caption{Ablations on the cross-modality lumbar spine segmentation (LSS) task (CT$\rightarrow$MRI). All results are given as mean over 3 random initializations. The best result per column is in bold.}
{
{
\begin{tabular}{c|cccccc}
    \toprule[0.8pt]
    \multirow{2}{*}{Method} & \multicolumn{6}{c}{Lumbar Spine CT $\rightarrow$ Lumbar Spine MRI}\\
    \cmidrule{2-7}
    & L1 & L2 & L3 & L4 & L5 & Avg \\
    \midrule[0.5pt]
    \textsc{Baseline} & 0.00 & 0.00 & 0.00 & 0.00 & 0.25 & 0.05 \\
    \midrule[0.5pt]
    \textsc{w/o Reg.} & 63.00 & 60.57 & 62.73 & 73.00 & 78.07 & 67.47 \\
    \textsc{w/ KL-Div} & 67.17 & 67.80 & 70.20 & 74.53 & 76.73 & 71.30 \\
    \textsc{w/ S. InfoNCE} & 64.80 & 66.63 & 68.40 & 73.90 & 78.07 & 70.37 \\
    \textsc{Ours} & \textbf{72.23} & \textbf{73.00} & \textbf{74.20} & \textbf{76.57} & \textbf{78.70} & \textbf{74.93} \\
    \bottomrule[0.8pt]
\end{tabular}
}
}
\label{tab:4}
\vspace{-0.3cm}
\end{table}

\begin{table}[ht]
\scriptsize
\centering
\caption{Ablations on the cross-modality lumbar spine segmentation (LSS) task (CT$\rightarrow$MRI). All results are given as mean over 3 random initializations. The best result per column is in bold.}
{
{
\begin{tabular}{c|cccccc}
    \toprule[0.8pt]
    \multirow{2}{*}{Method} & \multicolumn{6}{c}{Lumbar Spine CT $\rightarrow$ Lumbar Spine MRI}\\
    \cmidrule{2-7}
    & L1 & L2 & L3 & L4 & L5 & Avg \\
    \midrule[0.5pt]
    \textsc{Baseline} & 0.00 & 0.00 & 0.00 & 0.00 & 0.25 & 0.05 \\
    \midrule[0.5pt]
    \textsc{BigAug}~\cite{zhang2020generalizing} & 0.23 & 0.60 & 0.40 & 0.60 & 0.20 & 0.43 \\
    \textsc{GIN+IPA}~\cite{ouyang2022causality} & 71.47 & 72.70 & 72.93 & 74.63 & 74.40 & 73.23 \\
    \textsc{Ours} & \textbf{72.23} & \textbf{73.00} & \textbf{74.20} & \textbf{76.57} & \textbf{78.70} & \textbf{74.93} \\
    \bottomrule[0.8pt]
\end{tabular}
}
}
\label{tab:5}
\vspace{-0.3cm}
\end{table}

\begin{table}[ht]
\scriptsize
\centering
\caption{Ablations on the cross-modality lumbar spine segmentation (LSS) task (CT$\rightarrow$MRI). All results are given as mean over 3 random initializations. The best result per column is in bold.}
{
\begin{tabular}{c|cccccc}
    \toprule[0.8pt]
    \multirow{2}{*}{Method} & \multicolumn{6}{c}{Lumbar Spine CT $\rightarrow$ Lumbar Spine MRI}\\
    \cmidrule{2-7}
    & L1 & L2 & L3 & L4 & L5 & Avg \\
    \midrule[0.5pt]
    \textsc{Baseline} & 0.00 & 0.00 & 0.00 & 0.00 & 0.25 & 0.05 \\
    \midrule[0.5pt]
    \textsc{w/o Styel Swap} & 45.53 & 47.63 & 42.67 & 47.43 & 40.83 & 44.83 \\
    \textsc{Ours} & \textbf{72.23} & \textbf{73.00} & \textbf{74.20} & \textbf{76.57} & \textbf{78.70} & \textbf{74.93} \\
    \bottomrule[0.8pt]
\end{tabular}
}
\label{tab: 6}
\vspace{-0.3cm}
\end{table}

\noindent\textbf{Efficacy of Diffusion-based Style Intervention.}
We {conjecture} the comprehensive style intervention is crucial for extracting content features. To validate this, we compare our diffusion-based style intervention with two alternatives: (1) basic photometric and geometric augmentations~\cite{zhang2020generalizing} discussed in~\cref{exp_setting} (BigAug), and (2) the global intensity non-linear augmentation and interventional pseudo-correlation augmentation (GIN+IPA) method proposed in~\cite{ouyang2022causality}. Results in~\cref{tab:4} demonstrate that basic augmentations alone are insufficient for cross-modality style changes, with InfoNCE loss providing only a 0.38 \gls{pp} improvement in average Dice score. Our method surpasses the augmentation strategies proposed in CSDG~\cite{ouyang2022causality}, the second-best method for the \gls{lss} task (CT$\rightarrow$MRI), by 1.7 \gls{pp}. This improvement demonstrates the effectiveness of using a controlled \gls{sd} model for style intervention. 

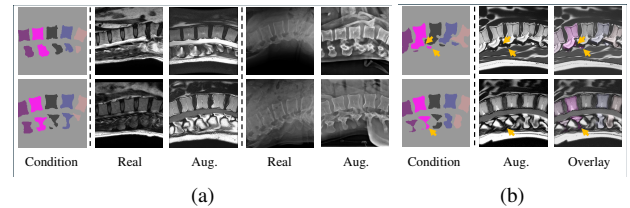
\begin{figure}[t]
    \centering
    \begin{subfigure}{0.603\linewidth}
        \centering
        \begin{overpic}[width=1 \columnwidth]{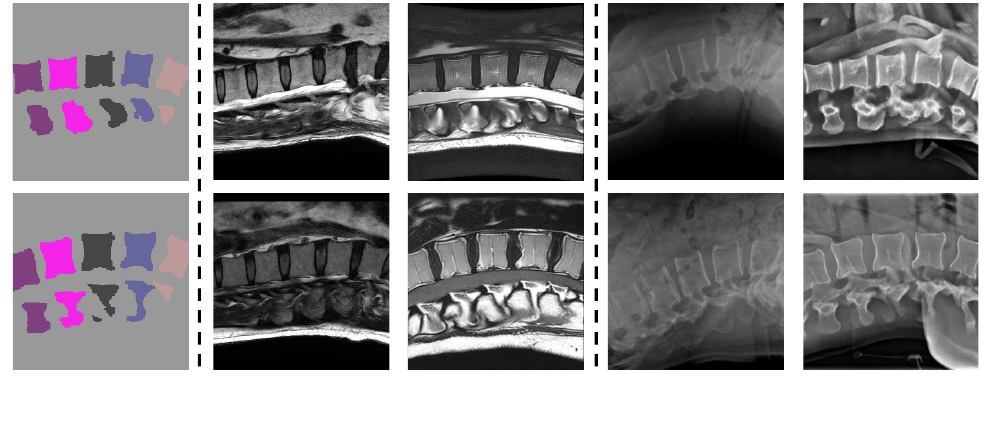}
            \put(3.0,3.0){\color{black}\tiny{Condition}}
            \put(27.5, 3.0){\color{black}\tiny{Real}}
            \put(47.0, 3.0){\color{black}\tiny{Aug.}}
            \put(67.1, 3.0){\color{black}\tiny{Real}}
            \put(87.1,3.0){\color{black}\tiny{Aug.}}
        \end{overpic}
        \caption{}
        \label{fig6_a}
    \end{subfigure}
    % \hfill
    \begin{subfigure}{0.363\linewidth}
        \centering
        \begin{overpic}[width=1 \columnwidth]{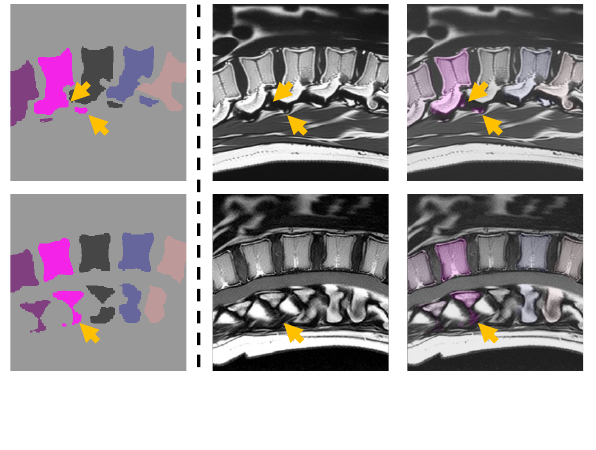}
            \put(4.0, 4.9){\color{black}\tiny{Condition}}
            \put(46.0, 4.9){\color{black}\tiny{Aug.}}
            \put(74.5, 4.9){\color{black}\tiny{Overlay}}
        \end{overpic}
        \caption{}
        \label{fig6_b}
    \end{subfigure}
    \caption{Visualization of diffusion-based style intervention for the LSS task. (a) Augmentation (Aug.) generated with style prompt \emph{"MRI"} and \emph{"X-Ray"}. (b) Examples showing limitations.}
    \label{fig6}
    % \vspace{-0.3cm}
\end{figure}

We also visualize our diffusion-generated augmentations in~\cref{fig6_a}, showcasing their close resemblance to target domain styles. The augmentations generated with the style prompt \emph{"MRI"} accurately replicate the high soft-tissue contrast characteristic of lumbar spine MRI, distinguishing vertebral bodies, intervertebral discs, and the spinal cord. The spinal cord's visibility, particularly its bright appearance on T2-weighted MRI, is difficult to simulate with generic image-level transformations or random intensity changes on CT scans~\cite{zhou2022dn, ouyang2022causality, su2023rethinking}. Conversely, the X-Ray augmentations effectively capture the high-contrast appearance and edge enhancement typical of radiographic imaging, particularly along vertebral body contours. However, our method inherits ControlNet's limitations in handling fine-grained structures. \cref{fig6_b} illustrates this for the \gls{lss} task (source domain CT), where generated augmentations satisfy conditions on large segmentation regions-of-interests (\ie, vertebral body) but struggle with smaller details like the posterior vertebral arch. This unintended alteration of content (\ie, segmentation masks) could have a potential negative impact on the generalization performance.

\noindent\textbf{Efficacy of Style Swap.}
To study the efficacy of the Style Swap technique~\cite{jia2023dginstyle}, we directly train an additional ControlNet with $U^B$, omitting the instance fine-tuning stage with DreamBooth~\cite{ruiz2023dreambooth}. Results in~\cref{tab: 6} show that bypassing the fine-tuning stage leads to average Dice score drop of $30.10$ \gls{pp}. These findings support our hypothesis that the style prior stored in $U^B$ may be compromised due to a mismatch between its prior information and the domain knowledge of $D_0$, which leads to insufficient simulation of cross-modal domain shifts and, consequently, worse generalization performance.

\section{Conclusion}
In this paper, we address \gls{sdg} for cross-modality medical image segmentation from a causal perspective and 
% demonstrate the effectiveness of contrastive learning in extracting domain-invariant content features. We 
introduce a style intervention method that simulates appearances from various imaging modalities utilizing large-scale pre-trained \gls{sd} model with ControlNet. Experiments on three segmentation tasks acorss different anatomies and imaging modalities show that our approach consistently outperforms state-of-the-art \gls{sdg} methods for medical imaging. 

However, our method inherits the limitations of the ControlNet, which struggles to generate images corresponding to fine-grained details in the conditional segmentation masks. Thus, a more refined control on anatomical structure such as~\cite{konz2024anatomically} is desirable for the exact correspondence between generated anatomical structures and given segmentation masks. Alternatively, we can incorporate uncertainty estimation methods~\cite{gal2016dropout, tagasovska2019single} into our framework to address the noisy labels resulting from the coarse control.

\section*{Acknowledgements}
We gratefully acknowledge funding from the ETH AI Center, the Swiss Federal Institutes of Technology strategic focus area of personalized health and related technologies project 2022-643, and the ETH core funding.

\clearpage
%%%%%%%%% REFERENCES
{\small
\bibliographystyle{ieee_fullname}
\bibliography{egbib}
}

\clearpage
%%%%%%%%% SUPPLEMENT
\setcounter{section}{0}
\renewcommand{\thesection}{\Alph{section}}
% WACV 2025 Paper Template
% based on the WACV 2024 template, which is
% based on the CVPR 2023 template (https://media.icml.cc/Conferences/CVPR2023/cvpr2023-author_kit-v1_1-1.zip) with 2-track changes from the WACV 2023 template (https://github.com/wacv-pcs/WACV-2023-Author-Kit)
% based on the CVPR template provided by Ming-Ming Cheng (https://github.com/MCG-NKU/CVPR_Template)
% modified and extended by Stefan Roth (stefan.roth@NOSPAMtu-darmstadt.de)

% Baselines:
\newcommand{\mixstyle}{\textsc{MixStyle}}
\newcommand{\randconv}{\textsc{RandConv}}
\newcommand{\dn}{\textsc{DualNorm}}
\newcommand{\csdg}{\textsc{CSDG}}
\newcommand{\slaug}{\textsc{SLAug}}

\section*{Appendix}

%%%%%%%%% BODY TEXT
\glsresetall
\noindent In this supplemental material, we include additional backgrounds, visualization results, and analyses. The contents of the individual sections are:
\begin{itemize}
    \item Appendix~\cref{app: theory}: Background on domain generalization and causality.
    \item Appendix~\cref{app: ablation}: visualization results and qualitative analysis of the ablation study on the impact of fine-tuning the SD U-Net on source domain $D_0$ using DreamBooth~\cite{ruiz2023dreambooth} before the ControlNet~\cite{zhang2023adding} training.
    \item Appendix~\cref{app: seg_viz}: Additional visualization results of segmentation maps predicted by our method and competing baselines on the \gls{as}, \gls{lss} and \gls{ls} tasks.
    \item Appendix~\cref{app: sty_vis}: Additional visualization results of diffusion-generated augmentations for the \gls{as}, \gls{lss} and \gls{ls} tasks with different style-intervention prompts.
    \item Appendix~\cref{app: data}: Dataset and preprocessing details.
    \item Appendix~\cref{app:implementation}: Implementation and training details of our method and baselines, including model architectures, hyperparameters, etc.
\end{itemize}

\section{Background: Generalization and Causality}
\label{app: theory}
Let $X \in \mathcal{X}$ and $Y \in \mathcal{Y}$ represent input images and corresponding segmentation masks. We further assume that an expert (or oracle) is able to provide correct segmentation masks $Y$ from observations $X$ alone. In the context of \gls{sdg}, we assume that we are given training pairs from a single domain $D_0$: $\{(X^0_i, Y^0_i)\}_{i=1}^{n_0}$, and a fixed set of target domains $D_1, \ldots, D_N$. Each dataset $D_e$ contains \gls{iid} samples from some probability distribution $P(X^e, Y^e)$. We aim to obtain an optimal predictor $f$ to enable \gls{ood} generalization. In particular, $f$ is trained on $D_0$ such that $f$ minimizes the worst case risk $R^e(f) := \mathbb{E}_{X^e, Y^e} \left[ \ell(f(X^e), Y^e) \right]$ for any given target domain $D_e$:
\begin{equation}\label{eq: risk}
\begin{split}
R_{\text{OOD}}(f) = \max_{e \in \{1, \ldots, N\}} R^e(f).
\end{split}
\end{equation}
Arjovsky et al.~\cite{arjovsky2019invariant} demonstrate that an invariant predictor would obtain an optimal solution for \cref{eq: risk}:
% \textbf{Definition 1\cite{arjovsky2019invariant}} A data representation \( \Phi : \mathcal{X} \rightarrow \mathcal{H} \) elicits an invariant predictor \( w \circ \Phi \) across environments \( \mathcal{E} \) if there is a classifier \( w : \mathcal{H} \rightarrow \mathcal{Y} \) simultaneously optimal for all environments, that is,
% \[
% w \in \arg \min_{\bar{w} : \mathcal{H} \rightarrow \mathcal{Y}} R_e(\bar{w} \circ \Phi) \quad \text{for all } e \in \{1, \ldots, N\}.
% \]
\begin{definition}[\cite{arjovsky2019invariant}]
\label{def-1}
\textnormal{A data representation \( \Phi : \mathcal{X} \rightarrow \mathcal{H} \) elicits an invariant predictor \( w \circ \Phi \) across environments (domains) \( \mathcal{E} \) if there is a classifier \( w : \mathcal{H} \rightarrow \mathcal{Y} \) simultaneously optimal for all environments, that is,
\[
w \in \arg \min_{\bar{w} : \mathcal{H} \rightarrow \mathcal{Y}} R^e(\bar{w} \circ \Phi) \quad \text{for all } e \in \mathcal{E}.
\]}
\end{definition}

Arjovsky et al.~\cite{arjovsky2019invariant} propose an \gls{irm} algorithm to obtain this invariant solution from \textit{multiple} domains and show that this solution is optimal if and only if it uses only the direct causal parents of $Y$ in the corresponding \gls{scm}~\cite{pearl2009causality}. In this work, we demonstrate how to achieve this optimal solution given a \emph{single} source domain. 

We extend common assumptions~\cite{lv2022causality, ouyang2022causality} on the data generative process from a causal perspective by allowing an additional causal relation from content to style as proposed in~\cite{von2021self}:
\begin{equation}\label{eq: SCM}
\begin{split}
C := g_C(U_c), \\
S := g_S(C, U_s), \\
X := g_X(C, S), \\
Y := g_Y(C),
\end{split}
\end{equation}
where $C$ denotes the latent content variable, $S$ denotes the latent style variable, $X \sim p_X$ is an observed input variable, $Y$ is the observed segmentation mask, $(U_c, U_s) \sim p_{u_c} \times p_{u_s}$ are independent exogenous variables, and $g_C$, $g_S$, $g_X$, $g_Y$ are deterministic functions. We assume that different domains are generated via an intervention on the style variable $S$. 

Under the SCM described above (\cref{eq: SCM}), von Kügelgen et al.~\cite{von2021self} theoretically prove that if the augmented pairs of views $(X, X^+)$ in contrastive methods are generated under the principle of a "soft" intervention on $S$, then the InfoNCE~\cite{oord2018representation} objective combined with an encoder function $\Phi$ (\eg, neural network) \textit{identifies} the invariant content $C$ partition in the generative process described above (\cref{eq: SCM}): 
\begin{equation}\label{eq: InfoNCE}
    \mathcal{L}_{\text{InfoNCE}} = -\mathbb{E}_{X \sim p_X} \left[ \log \frac{\exp(\text{sim}(\nu, \nu^+) / \tau)}{\sum_{X^- \in \mathcal{N}} \exp(\text{sim}(\nu, \nu^-) / \tau)} \right],
\end{equation}
where $\nu = \Phi(X)$, $\text{sim}(\cdot,\cdot)$ is some similarity metric (\eg, cosine similarity), $\tau$ is a temperature parameter, and $X^-$ are negative pairs.

In real-world applications, direct intervention on the style variable is infeasible as it is unobserved. For instance, in medical imaging, this would require scanning the same patient using different imaging modalities multiple times under controlled conditions—a process that is often impractical for acquiring training data. However, Ilse et al.~\cite{ilse2021selecting} introduce the concept of \textit{intervention-augmentation equivariance}, formally demonstrating that data augmentation can serve as a surrogate tool for simulating interventions:
% In real-world applications, we cannot perform a direct intervention on the style variable as it is unobserved. For example, in medical imageing setting, it would mean that we need to scan the same patient using different imaging modalities multiple times in a controlled environment. It is often impractical to aquire such training data. However, Ilse et al. introduced a concept of \textit{intervention-augmentation equivariance} and formally demonstrated that such data augmentation can serve as a surrogate tool for simulating interventions\cite{ilse2021selecting}:

% \textbf{Definition 2\cite{ilse2021selecting}} The causal process $g_x$, is \emph{intervention-augmentation equivariant} if for every considered stochastic data augmentation transformation $\text{aug}(\cdot)$ on $x \in \mathcal{X}$ we have a corresponding noise intervention $do(\cdot)$ on $S$ such that:
% \begin{equation}
%     \text{aug}(x) = g_x(do(S = s), c).
%     % x_\text{aug} = \text{aug}(x) = \text{aug}(g_x(c, s)).
% \end{equation}
\begin{definition}[\cite{ilse2021selecting}]
\label{def-2}
\textnormal{The causal process \( g_X \) is \emph{intervention-augmentation equivariant} if for every considered stochastic data augmentation transformation \(\text{aug}(\cdot)\) on \( X \in \mathcal{X} \) we have a corresponding noise intervention \( do(\cdot) \) on \( S \) such that:
\begin{equation}
    \text{aug}(X) = g_X(do(S = s), c).
    % x_\text{aug} = \text{aug}(x) = \text{aug}(g_x(c, s)).
\end{equation}}
\end{definition}

\section{Visualizations of Ablation Study}
\label{app: ablation}
We conduct an ablation study on the \gls{lss} task (CT$\rightarrow$MRI) to investigate the efficacy of the Style Swap technique~\cite{jia2023dginstyle} in the main paper. This technique involves two steps: first, fine-tuning the pretrained \gls{sd} U-Net ($U^B$) on the source domain using DreamBooth~\cite{ruiz2023dreambooth} to obtain an instance-tuned \gls{sd} U-Net $U^{D_0}$, and then train the ControlNet~\cite{zhang2023adding} with $U^{D_0}$ so that it can focus on learning to inject image conditions into the generation process rather than domain information from $D_0$. Specifically, we compare this method to directly training the ControlNet with $U^B$, omitting the instance fine-tuning stage with DreamBooth. 
% Results in~\cref{tab:5} demonstrate that bypassing the fine-tuning stage leads to a significant performance drop in the target domain, with the average Dice score decreasing by $30.10$ \gls{pp}. These findings support our hypothesis that the style prior stored in $U^B$ may be compromised due to a mismatch between its prior information and the domain knowledge of $D_0$ used to train the ControlNet.

\begin{figure}[t]
\centering
\includegraphics[width=1.0\columnwidth]{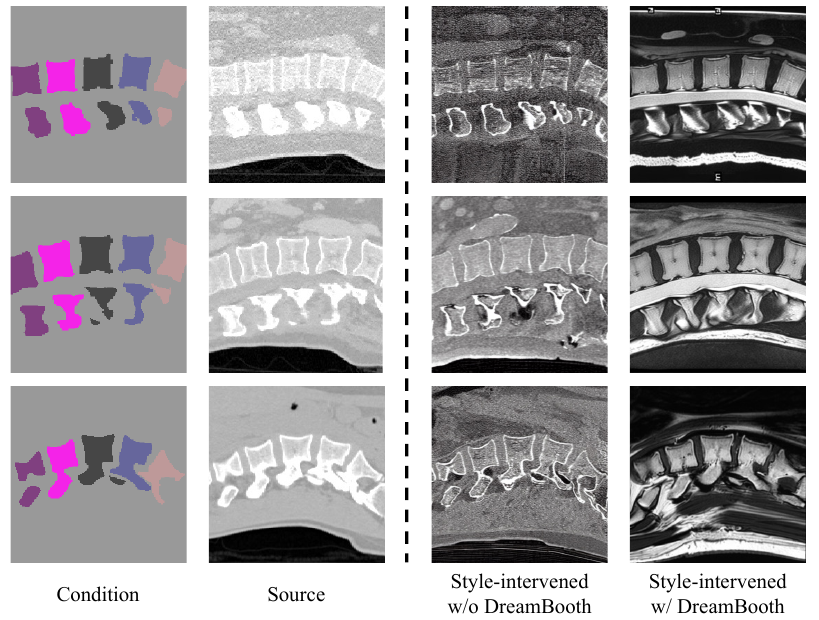} 
\caption{Visualization results on diffusion-based style intervention for \gls{lss} task (CT$\rightarrow$MRI). The first and second column show the image conditions (\ie, segmentation masks) and the corresponding source images, respectively. The third column shows the style-intervened images using the ControlNet trained direcly with pretrained \gls{sd} U-Net (\ie, no instance fine-tuning stage using DreamBooth). The forth column shows the style-intervened images using the ControlNet trained with instance-fine-tuned U-Net. Both of them are generated using the style-intervention prompt "\emph{sagittal lumbar spine MRI}".}
\label{fig: supp-1}
\vspace{-10pt}
\end{figure}

To further illustrate this point, we visualize the images generated by $U^B$ using two versions of ControlNet: one trained with $U^{D_0}$ (fine-tuned on $D_0$ with DreamBooth) and another trained with $U^B$ from the original \gls{sd} model, using the same style-intervention prompt. As shown in~\cref{fig: supp-1}, images generated using ControlNet trained directly with $U^B$ on $D_0$ (top row) with the prompt "\emph{sagittal lumbar spine MRI}" retain characteristics of CT modality rather than MRI. However, these images appear darker than typical CT images, which is more characteristic of MRI, indicating partial intervention on the style variable.

Conversely, images generated using the ControlNet trained with instance-fine-tuned $U^{D_0}$ (bottom row) exhibit the appearance of MRI images rather than CT, demonstrating the effectiveness of style variable intervention through prompting when using the Style Swap technique.

% \begin{table}[ht]
% \scriptsize
% \centering
% \caption{Ablations on the cross-modality lumbar spine segmentation (LSS) task (CT$\rightarrow$MRI). All results are given as mean over 3 random initializations. The best result per column is in bold.}
% {
% {
% \begin{tabular}{c|cccccc}
%     \toprule[0.8pt]
%     \multirow{2}{*}{Method} & \multicolumn{6}{c}{Lumbar Spine CT $\rightarrow$ Lumbar Spine MRI}\\
%     \cmidrule{2-7}
%     & L1 & L2 & L3 & L4 & L5 & Avg \\
%     \midrule[0.5pt]
%     \textsc{Baseline} & 0.00 & 0.00 & 0.00 & 0.00 & 0.25 & 0.05 \\
%     \midrule[0.5pt]
%     \textsc{w/o DreamBooth} & 45.53 & 47.63 & 42.67 & 47.43 & 40.83 & 44.83 \\
%     \textsc{Ours} & \textbf{72.23} & \textbf{73.00} & \textbf{74.20} & \textbf{76.57} & \textbf{78.70} & \textbf{74.93} \\
%     \bottomrule[0.8pt]
% \end{tabular}
% }
% }
% \label{tab:5}
% \vspace{-0.3cm}
% \end{table}

\section{Visualizations of Predicted Segmentation Masks}
\label{app: seg_viz}
\begin{figure*}[t]
\centering
\begin{overpic}[width=0.9 \textwidth]{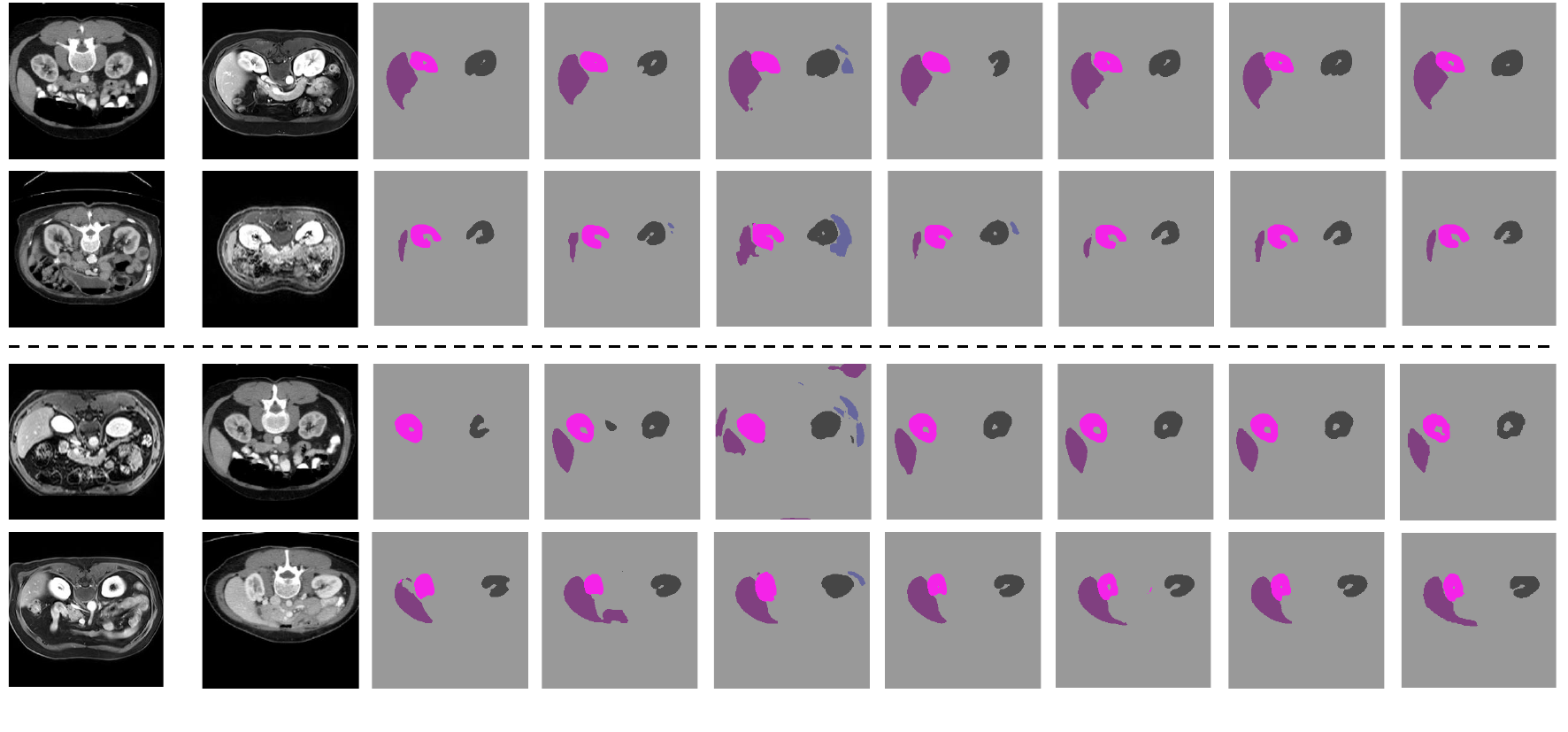}
    \put(0.5, 1.2){\color{black}\scriptsize{Source Domain}}
    \put(13.75, 1.2){\color{black}\scriptsize{Target Image}}
    \put(26.05, 1.2){\color{black}\scriptsize{MixStyle}}
    \put(36.6, 1.2){\color{black}\scriptsize{RandConv}}
    \put(47.4, 1.2){\color{black}\scriptsize{DualNorm}}
    \put(59.2, 1.2){\color{black}\scriptsize{CSDG}}
    \put(70.3, 1.2){\color{black}\scriptsize{SLAug}}
    \put(81.9, 1.2){\color{black}\scriptsize{Ours}}
    \put(90.2, 1.2){\color{black}\scriptsize{Ground Truth}}
\end{overpic}
\caption{Additional visualization results on \gls{as} task of different methods. First two rows: “CT to MRI” task; Last two rows: “MRI to CT” task.}
\label{fig: supp-2}
\end{figure*}

\begin{figure*}[t]
\centering
\begin{overpic}[width=0.9 \textwidth]{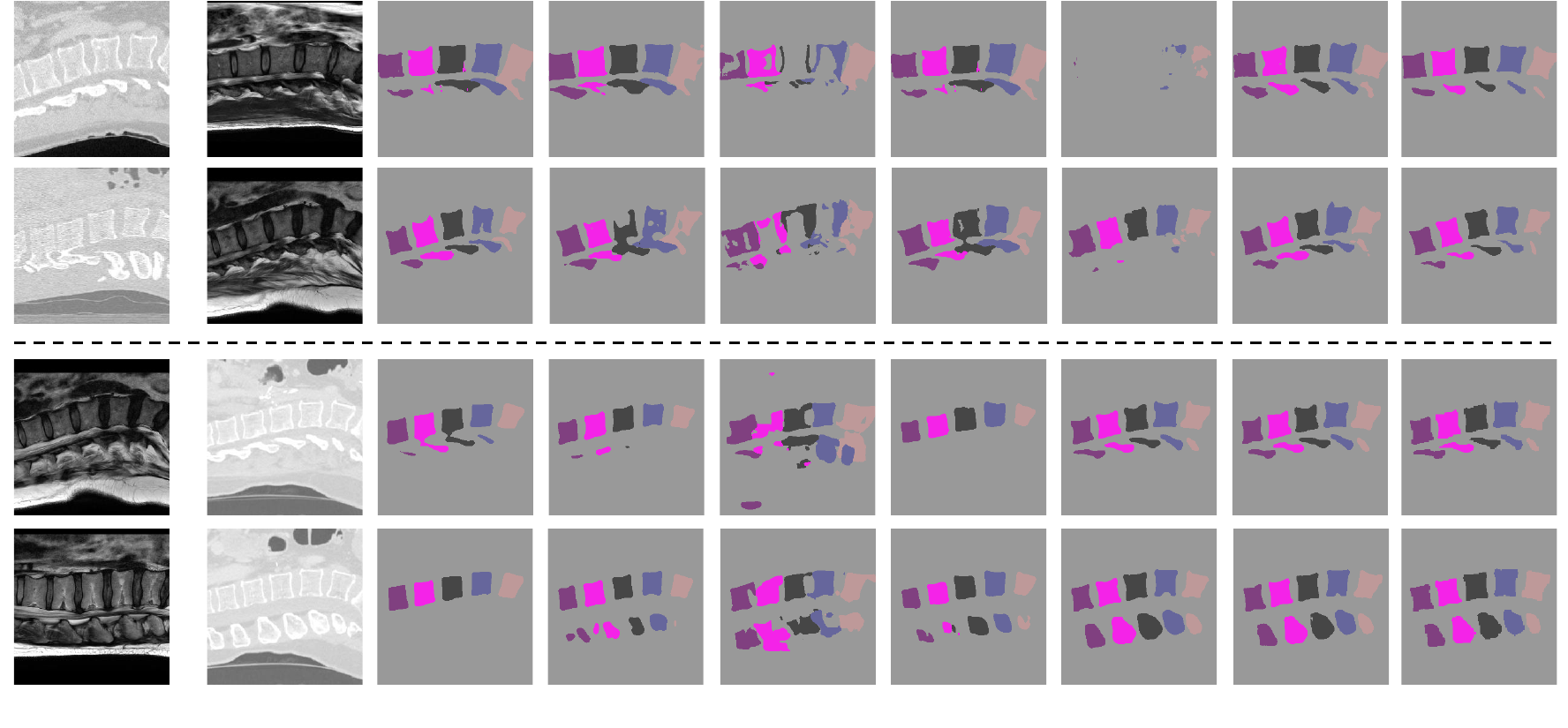}
    \put(0.65, 1.2){\color{black}\scriptsize{Source Domain}}
    \put(13.75, 1.2){\color{black}\scriptsize{Target Image}}
    \put(26.05, 1.2){\color{black}\scriptsize{MixStyle}}
    \put(36.6, 1.2){\color{black}\scriptsize{RandConv}}
    \put(47.4, 1.2){\color{black}\scriptsize{DualNorm}}
    \put(59.3, 1.2){\color{black}\scriptsize{CSDG}}
    \put(70.3, 1.2){\color{black}\scriptsize{SLAug}}
    \put(81.9, 1.2){\color{black}\scriptsize{Ours}}
    \put(90.2, 1.2){\color{black}\scriptsize{Ground Truth}}
\end{overpic}
\caption{Additional visualization results on \gls{lss} task of different methods. First two rows: “CT to MRI” task; Last two rows: “MRI to CT” task.}
\label{fig: supp-3}
\vspace{-0.3cm}
\end{figure*}

\begin{figure*}[t]
\centering
\includegraphics[width=0.9 \textwidth]{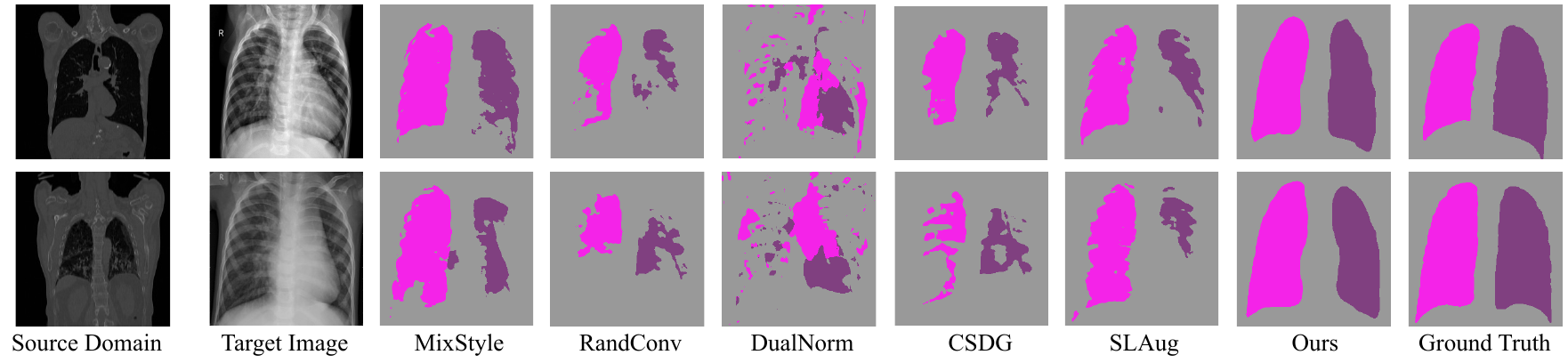}
\caption{Additional visualization results on \gls{ls} task of different methods. First two rows: “CT to X-Ray” task.}
\label{fig: supp-4}
\vspace{-0.3cm}
\end{figure*}

We present additional visualizations of segmentation results from our method and other \gls{dg} approaches for the \gls{as}, \gls{lss} and \gls{ls} tasks in~\cref{fig: supp-2},~\cref{fig: supp-3} and~\cref{fig: supp-4}, respectively. The first two columns display source and target domain images, illustrating the domain shift. Consistent with our main findings, these visualizations demonstrate that our method consistently produces superior segmentation masks in the unseen target domain. Notably, the generated masks exhibit not only higher accuracy but also enhanced spatial continuity of the foreground classes.

\section{Visualizations of Diffusion-generated Augmentations}
\label{app: sty_vis}
\begin{figure*}[t]
\centering
\includegraphics[width=0.9 \textwidth]{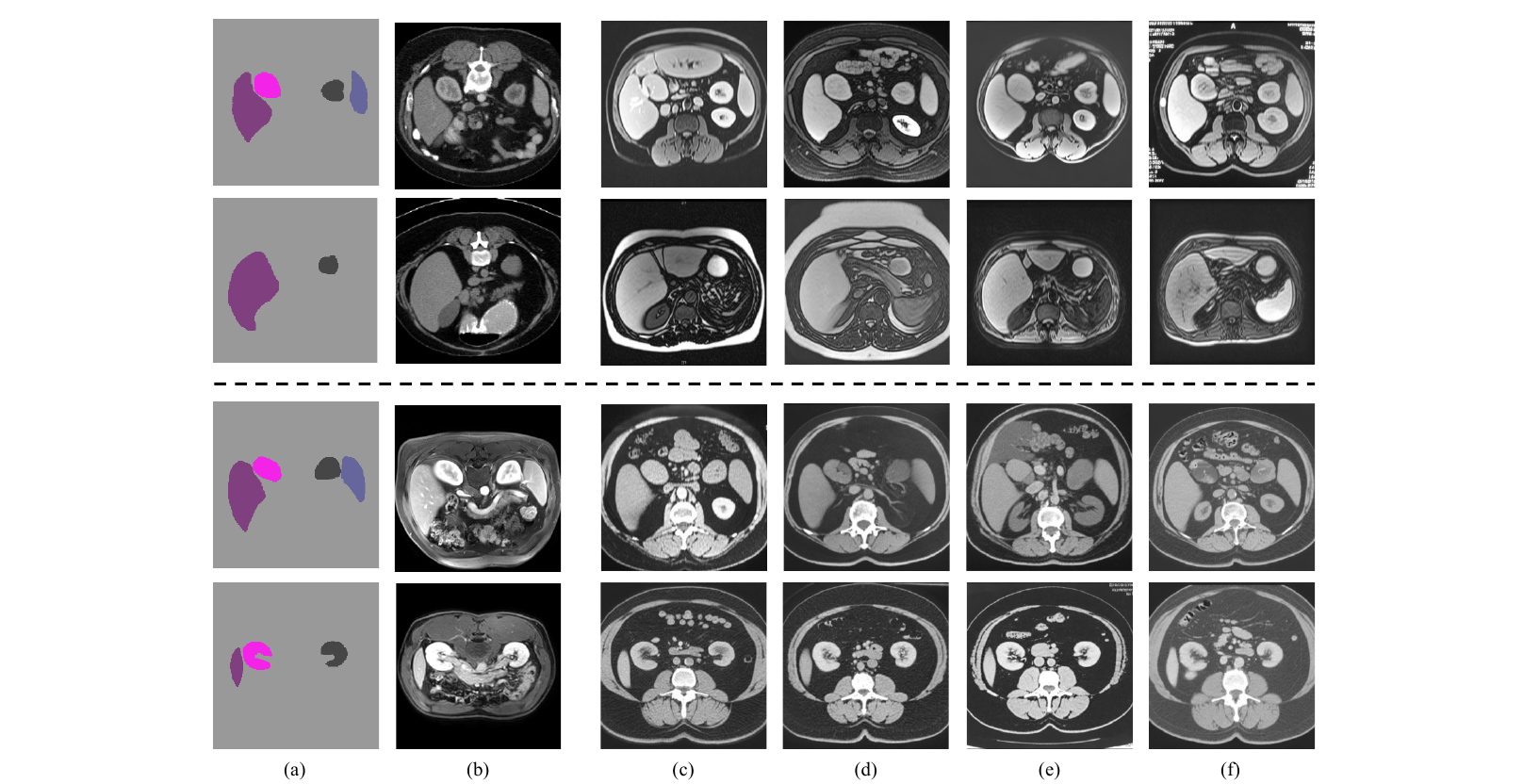}
\caption{
Visualization results of diffusion-based style intervention for the \gls{as} task. The top two rows show the "CT to MRI" task, while the bottom two rows display the "MRI to CT" task. The left columns present (a) the image conditions (\ie, segmentation masks) and (b) the corresponding source images. The right columns (c-f) show the style-intervened images generated using specific style-intervention prompts. For the first row: "\emph{axial liver left kidney right kidney spleen MRI}"; second row: "\emph{axial liver spleen MRI}"; third row: "\emph{axial liver left kidney right kidney spleen CT}"; and fourth row: "\emph{axial liver left kidney right kidney CT}". 
}
\label{fig: supp-5}
\vspace{-0.3cm}
\end{figure*}

\begin{figure*}[t]
\centering
\includegraphics[width=0.9 \textwidth]{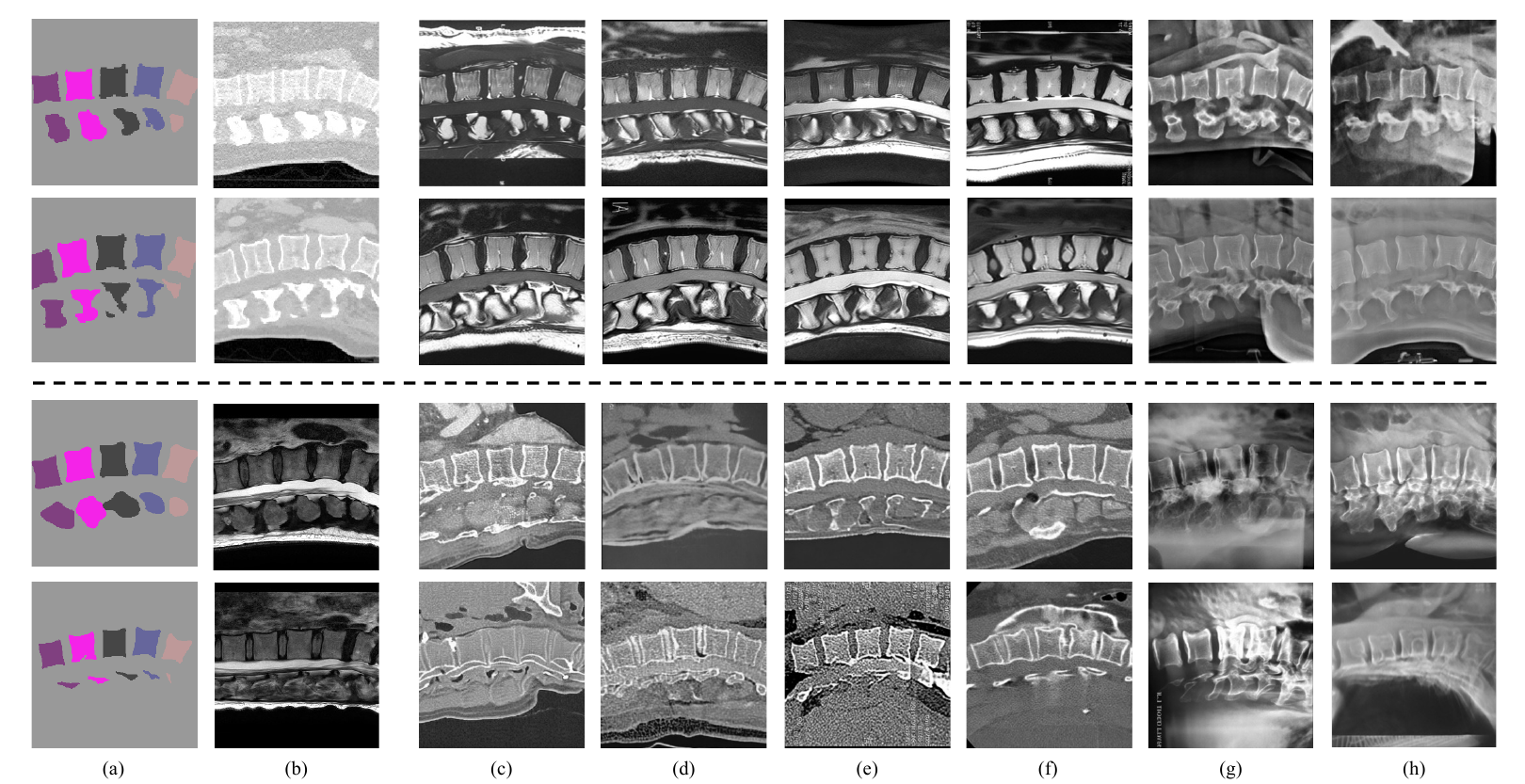}
\caption{
Visualization results of diffusion-based style intervention for the \gls{lss} task. The first two rows demonstrate the "CT to MRI" task, while the last two rows show the "MRI to CT" task. Columns on the left display (a) the image conditions (i.e., segmentation masks) and (b) the corresponding source images, respectively. For the first two rows (CT to MRI task), the right columns show style-intervened images generated using the following prompts: (c-d) "\emph{sagittal lumbar spine T1-MRI}"; (e-f) "\emph{sagittal lumbar spine T2-MRI}"; and (g-h) "\emph{sagittal lumbar spine X-ray}". For the last two rows (MRI to CT task), the right columns present style-intervened images generated using these prompts: (c-f) "\emph{sagittal lumbar spine CT}"; and (g-h) "\emph{sagittal lumbar spine X-ray}".
}
\label{fig: supp-6}
\vspace{-0.3cm}
\end{figure*}

\begin{figure*}[t]
\centering
\includegraphics[width=0.66 \textwidth]{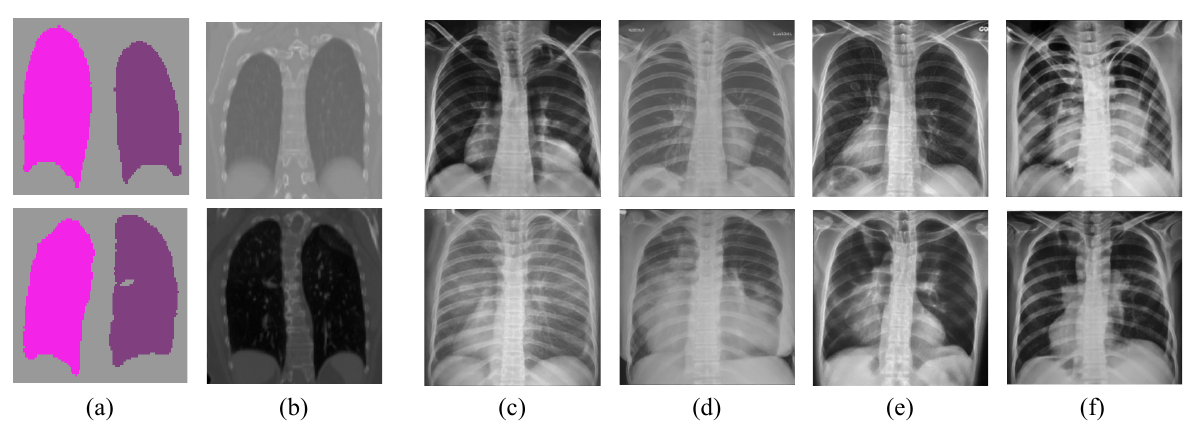}
\caption{
Visualization results of diffusion-based style intervention for the \gls{ls} task. The top two rows show the "CT to X-Ray" task. The left columns present (a) the image conditions (\ie, segmentation masks) and (b) the corresponding source images. The right columns (c-f) show the style-intervened images generated using style-intervention prompt:"\emph{chest CT left lung right lung}". 
}
\label{fig: supp-7}
\vspace{-0.3cm}
\end{figure*}

To demonstrate the effectiveness of our diffusion-based style intervention, we provide visualizations of images generated by our controlled diffusion model for the \gls{as},~\gls{lss} and~\gls{ls} tasks in~\cref{fig: supp-5},~\cref{fig: supp-6} and~\cref{fig: supp-7}, respectively. These visualizations illustrate that our diffusion-based augmentation strategy successfully leverages the rich generative prior of the \gls{sd} model, which has been trained on a large-scale medical imaging dataset comprising scans from diverse anatomies and imaging modalities. Our method performs comprehensive intervention on the style variable while preserving the content. These strong augmentations are crucial for extracting content features using contrastive learning. 

\section{Dataset And Preprocessing Details}
\label{app: data}
We evaluate our method on three cross-modality segmentation tasks: \gls{as}, \gls{lss} and \gls{ls} tasks. For the \gls{as} task, we have 20/10 volumes for train/test CT data and 40/20 volumes for train/test MRI data. For the \gls{lss} task, we randomly split all datasets into approximately 90\% train / 10\% test, resulting in 117/15 train/test volumes for CT , 182/20 volumes for MRI and 350/50 images for X-Ray. For the \gls{ls} task, we use the split proposed in the original paper~\cite{yang2017data} for CT data, which contains 36 volumes for training and 24 volumes for testing. For X-Ray scans, we first perform quality control (QT) on both images and labels. We filter out the cases where images are not X-Ray scans (\eg, some scans are axial slices of CT scans) and/or labels contains no foreground or a rectangle foreground. After the QT, we randomly split the dataset into approximately 90\% train / 10\% test, resulting in 5409/601 train/test images. We adopt the preprocessing steps from~\cite{ouyang2022causality} for all 3D volume data.

The common augmentations applied to all methods include affine transformations, elastic deformations, brightness and contrast adjustments, gamma corrections, and additive Gaussian noise.

\section{Implementation Details} 
\label{app:implementation}

\subsection{Loss Hyperparameters}
The hyperparameters for weighting different losses across all methods are presented in~\cref{tab:hyperparams}. All the parameters are selected based on a grid search over the range indicated in the "\emph{values}" column.

\begin{table}[ht]
\scriptsize
\centering
    \caption{Summary of hyperparameter optimization ranges for each method.}
    \label{tab:hyperparams}
    \centering
    \begin{tabular}{llll}
        \toprule
         Method & Parameter & Values & Description \\
         \midrule
         \multirow{3}{*}{Ours} & $\lambda_\mathrm{src}$ & $\{0.1, 1\}$ & \makecell[l]{Source image seg. loss} \\
         & $\lambda_\mathrm{sty}$ & $\{0.1, 1\}$ & \makecell[l]{Style-intervened image \\ seg. loss} \\
         & $\lambda_{contrast}$ & $\{1, 10\}$ & InforNCE regularization \\
         \midrule
         \multirow{2}{*}{\makecell[l]{RandConv~\cite{xu2020robust}}} & $\lambda_\mathrm{src}$ & $\{0.1, 0.5, 1\}$ & \makecell[l]{Source image seg. loss} \\
         & $\lambda_\mathrm{aug}$ & $\{0.1, 0.5, 1\}$ & {Augmented image seg. loss} \\
         \midrule
         \multirow{3}{*}{\makecell[l]{CSDG~\cite{ouyang2022causality}}} & $\lambda_\mathrm{src}$ & $\{0.1, 0.5, 1\}$ & \makecell[l]{Source image seg. loss} \\
         & $\lambda_\mathrm{aug}$ & $\{0.1, 0.5, 1\}$ & {Augmented image seg. loss} \\
         & $\lambda_{kl}$ & $\{1, 10\}$ & KL-DIV regularization \\
         \midrule
         \multirow{2}{*}{\makecell[l]{SLAug~\cite{su2023rethinking}}} & $\lambda_\mathrm{gla}$ & $\{0.1, 0.5, 1\}$ & \makecell[l]{Global location-scale\\ augmentation seg. loss} \\
         & $\lambda_\mathrm{sbf}$ & $\{0.1, 0.5, 1\}$ & \makecell[l]{Saliency-balancing fused\\ augmentation seg. loss} \\
        \bottomrule
    \end{tabular}
    \vspace{-0.5em}
\end{table}
\subsection{Baselines}
\textbf{$\mixstyle$:}
We implement MixStyle~\cite{zhou2021mixstyle} using the authors' provided code\footnote{\url{https://github.com/KaiyangZhou/mixstyle-release}}. To accommodate MixStyle's requirement for multiple source domains, we employ our diffusion-based augmentation to synthesize novel training domains with varied appearances across imaging modalities. We adopt the hyperparameters from the original implementation: mixing probability $p=0.5$, Beta distribution parameter $\alpha=0.1$, and "random" mixing method. Style mixing is applied twice after the first and second double-convolution blocks of the U-Net~\cite{ronneberger2015u} encoder.
% We implement MixStyle~\cite{zhou2021mixstyle} with the code\footnote{\url{https://github.com/KaiyangZhou/mixstyle-release}} provided by the authors. Since MixStyle requires probabilistic mixing of instance-level feature statistics of training samples from multiple source domains, we artificially create multiple source domains using the diffusion-based augmentation which synthesizes novel training domains with different appearances across diverse imaging modalities. We adopt the same hyperparameters in the released code. Specifically, we set the probability $p$ of using MixStyle to $0.5$, the parameter of the Beta distribution $\alpha=0.1$, and the mixing method to "\emph{random}". The style mixing is applied twice after the first and second double-convolution block of the U-Net~\cite{ronneberger2015u} encoder. 

\textbf{$\dn$:}
We use the original implementation of Dual-Normalization~\cite{zhou2022dn}\footnote{\url{https://github.com/zzzqzhou/Dual-Normalization}}. For each training image, we generate two source-similar and three source-dissimilar augmentations using nonlinear Bézier transformations. We use the default control points to define the Bézier curves for both source-similar and source-dissimilar augmentations, with 1000 time steps in both cases.

% We adopt the original implementation\footnote{\url{https://github.com/zzzqzhou/Dual-Normalization}} of Dual-Normalization~\cite{zhou2022dn}. For each training image, we generate two source-similar and three source-dissimilar augmentations with nonlinear Bézier transformation. Specifically, we use $((-1, -1), (-0.5, 0.5), (0.5, -0.5), (1, 1))$ and $((-1, -1), (-0.75, 0.75), (0.75, -0.75), (1, 1))$ as control points to define the Bézier curve to for the source-similar augmentation, and $((-1, -1), (-1, -1), (1, 1), (1, 1))$, $((-1, -1), (-0.5, 0.5), (0.5, -0.5), (1, 1))$ and $((-1, -1), (-0.75, 0.75), (0.75, -0.75), (1, 1))$ as control points for the source-dissimilar augmentation. We set the number of time steps to $1000$ in both cases. 

\textbf{$\csdg$:}
We adopt the model architecture from the original CSDG implementation~\cite{ouyang2022causality}\footnote{\url{https://github.com/cheng-01037/Causality-Medical-Image-Domain-Generalization}}. For the \gls{gin} module, we use 4 convolutional layers and 2 intermediate layers, with "Frobenius" normalization after the final layer. The \gls{ipa} module uses blending parameters $\epsilon=0.3$ and $\xi=1e^{-6}$, with control point spacing of 32, downsample scale of 2, and interpolation order of 2. Control point parameters are initialized using a Gaussian distribution.
% We adopt the model architecture from the original implementation\footnote{\url{https://github.com/cheng-01037/Causality-Medical-Image-Domain-Generalization}} of CSDG~\cite{ouyang2022causality}. For the \gls{gin} module, we set the number of convolutional layers to $4$, and the number of the intermediate layers to $2$. The output image is re-normalized uisng the "\emph{Frobenius}" normalization following the last convolutional layer. For the \gls{ipa} moduel, we set the blending parameters $\epsilon=0.3$ and $\xi=1e^{-6}$. The spacing between control points, downsample scale and interpolation order are set to $32$, $2$, $2$, respectively. We use Gaussian distribution to initialize control points parameters for optimization. 

\textbf{$\randconv$:}
We implement RandConv~\cite{xu2020robust} based on the CSDG implementation~\cite{ouyang2022causality}, removing the \gls{ipa} module to generate augmentations using only randomly-initialized convolutional layers. We adopt the same \gls{gin} module hyperparameters as CSDG.
% We implement RandConv based on the the implementation\footnote{\url{https://github.com/cheng-01037/Causality-Medical-Image-Domain-Generalization}} of $\csdg$~\cite{ouyang2022causality}. Specifically, we remove the \gls{ipa} module so that the augmentations are generated by only the randomly-initialized convolutional layers. We adopt the hyperparameters in the \gls{gin} module as GSDG. 

\textbf{$\slaug$:}
We use the model architecture and augmentation from the original SLAug implementation~\cite{su2023rethinking}\footnote{\url{https://github.com/Kaiseem/SLAug?tab=readme-ov-file}}. For the non-linear Bézier transformation, we set the background threshold to 0.01 and use 100,000 time steps. We initialize 4 control points based on the input image's intensity values, and gradually add another two points by uniformly sample a random value between the first and last elements of the current point array. The probability of random inversion is set to 0.5. For saliency-based fusion, we use a 2D B-spline kernel with interpolation order 2 and grid size 3.
% We adopt the model architecture and augmentation from the original implementation\footnote{\url{https://github.com/Kaiseem/SLAug?tab=readme-ov-file}} of SLAug~\cite{su2023rethinking}. For the non-linear Bézier transformation, we set the background threshold to $0.01$ and the number of time steps to $100000$. The number of the control point is set to $4$. We first initialize two control points using the maximum and minimum intensity value of the input image, then add another two points gradually by uniformly sample a random value between the first and last elements of the current point array. We also use a random inversion of the point array at a probability of $0.5$. For the saliency-based fusion, we adopt a 2D B-spline kernel with an order of interpolation being $2$ and a grid size of $3$, which is used to set the spacing between control points along height and width dimension of the image gradient.

\subsection{Diffusion-based Augmentation}
We implement DreamBooth~\cite{ruiz2023dreambooth} and ControlNet~\cite{zhang2023adding} using the HuggingFace Diffusers library\footnote{\url{https://github.com/huggingface/diffusers}}. During the sampling process, we adopt the UniPC sampler to accelerate the generation with 20 steps\cite{zhao2024unipc}. 

\end{document}